\documentclass[10pt,twocolumn]{article}

\usepackage[preprint]{antintlpaper}

\usepackage{multirow}
\usepackage{placeins}

\providecommand{\nolinenumbers}{}
\providecommand{\linenumbers}{}

\newcommand{\FACA}{\textsc{Faca}}
\newcommand{\Ao}{A^{\mathrm{o}}}
\newcommand{\Ap}{A^{\mathrm{p}}}

\AntTitle{Towards Better Agents for Multi-Turn User Interaction: The Next User Turn Is More Than Context}
\AntRunningTitle{The Next User Turn Is More Than Context}
\AntAuthors{Yiwen Zhao$^{1,2}$\AntEqualContributor \and Zhihao Wen$^{2}$\AntEqualContributor \and Yuchen Mao$^{2}$\AntEqualContributor \and Mingxuan Jiang$^{1}$ \and Yihao Hu$^{2}$ \and Pan Wang$^{2}$ \and Xin Zhang$^{2}$ \and Wei Wu$^{2}$}
\AntAffiliations{$^{1}$Fudan University \qquad $^{2}$Ant International, Ant Group}
\AntContact{Correspondence: z.wen@antgroup.com}
\AntEqualContributions
\AntDate{\today}
\AntLinks{}
\AntKeywords{LLM agents, multi-turn interaction, reinforcement learning, credit assignment, user simulation}
\AntAbstract{%
User-facing tool agents must coordinate dialogue and tool use as user goals
unfold over multiple turns. Yet interactive reinforcement learning typically
reduces each rollout to a terminal reward, assigning the same credit to
effective elicitation, errors, and later repair. The next user turn is more
than context: it also provides noisy, temporally local evidence about the
preceding user-to-user segment. We introduce \textbf{F}eedback-\textbf{A}ware \textbf{C}redit
\textbf{A}ssignment (\FACA{}), which aligns each reaction with that segment,
derives a locally normalized reaction advantage, and adds it to verified
terminal outcome advantage without an extra critic or rollout. Against an
outcome-only Interactive GRPO control matched in simulator, visible dialogue,
initialization, rollout, and optimization, \FACA{} improves the nine-domain
$\tau$-family average across three independently trained runs by 5.91 and 10.22
percentage points at 8B and 14B, respectively. Gains concentrate in Telecom;
at 8B, randomizing reaction polarity removes the Telecom gain. The same
ordering holds zero-shot on Pare-Bench and Co-Gym. These results demonstrate
that next-turn user reactions provide actionable local credit for improving
multi-turn user-interacting agents.
}

\hypersetup{pdfauthor={Yiwen Zhao, Zhihao Wen, Yuchen Mao, Mingxuan Jiang,
  Yihao Hu, Pan Wang, Xin Zhang, Wei Wu}}

\begin{document}

\makeanttitle

\section{Introduction}
\label{sec:intro}

\begin{figure}[!t]
    \centering
    \includegraphics[width=\columnwidth]{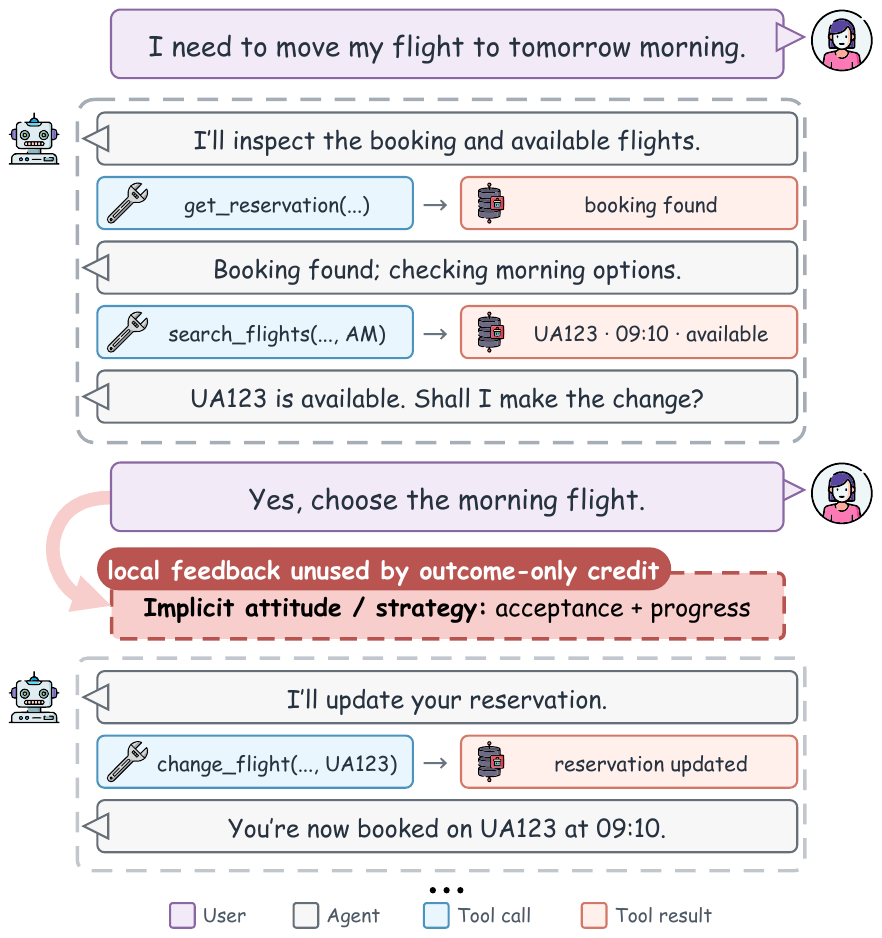}
    \caption{Multi-turn tool interaction. A user-to-user segment contains
    all agent responses, tool calls, and tool results between two adjacent user
    turns; the ellipsis denotes continued interaction.}
    \label{fig:intro-u2u}
\end{figure}

Tool-using language agents are moving from single-turn assistants toward
user-facing systems that operate reliably and safely over extended conversations. A travel agent
may need to inspect a reservation, elicit missing preferences, compare options,
obtain authorization, and only then modify the database. Unlike fully specified tasks
that provide all relevant information upfront, these interactions require the
agent to discover user intent while acting in an external environment under
evolving constraints. Goals may be omitted initially or revealed only after the
agent elicits them, making the user an active participant rather than a fixed input
\citep{yao2024taubench,qian2025userbench,barres2025tau2bench}.


Figure~\ref{fig:intro-u2u} illustrates this structure. Between adjacent user
turns \(u_t\) and \(u_{t+1}\), the agent may produce several messages and tool
interactions before returning control. We call this block a
\emph{user-to-user (U2U) segment}. The next user turn may provide information,
approve an action, reject a proposal, or correct a misunderstanding. Success
therefore requires coordinating dialogue, tool use, and user decisions across
segments throughout the evolving interaction process.

Interactive benchmarks in the $\tau$ family expose these demands
\citep{yao2024taubench,barres2025tau2bench,shi2026tauknowledge}. Training
frameworks such as MUA-RL and UserRL go further by placing an LLM-simulated user
inside reinforcement-learning rollouts
\citep{zhao2025muarl,qian2025userrl}. This shift lets the policy explore complete
conversations while the final database or world state provides objective
success supervision. Outcome-only credit accommodates many valid trajectories,
but collapses their internal structure: effective elicitation, errors, and
later repair receive the same advantage. The reward says \emph{whether} a
trajectory succeeded, but not \emph{where} it changed course.

Our central observation is that \textbf{the next user turn is more than context}. Prospectively, it guides the next decision; retrospectively, it provides local evidence about the preceding U2U segment. Supplying requested
information can indicate effective elicitation; challenges or clarifications can expose friction. These context-dependent reactions are not correctness labels, yet outcome-only training leaves them unused. Can this process evidence improve local credit while verified task completion remains the outcome objective?

We introduce Feedback-Aware Credit
Assignment (\FACA{}). \FACA{} aligns
each next-user reaction with the preceding U2U segment, normalizes reactions within a local rollout group, and adds the resulting process advantage to terminal outcome advantage. \textbf{It changes only credit assignment}: \FACA{} and the
matched outcome-only Interactive GRPO control share the frozen simulator, agent-visible dialogue, initialization, rollout, and optimization, while private reaction metadata remains hidden from the agent.

Across three independently trained runs per scale, \FACA{} improves the
nine-domain \(\tau\)-family average by \textbf{5.91} and \textbf{10.22}
percentage points at 8B and 14B, respectively. Effects are heterogeneous, with the largest gains in
Telecom; at 8B, randomizing reaction polarity removes the Telecom gain.
\FACA{}-trained policies also outperform matched controls zero-shot on
Pare-Bench and Co-Gym. These results support a conditional benefit when user
reactions carry informative local structure, rather than universal agent
improvement.

Overall, our main contributions are as follows:


\begin{itemize}[leftmargin=*]
    \item \textbf{Formulation.} We formalize U2U segments as local credit-assignment units, and subsequent user reactions as temporally localized evidence of interaction progress.
    \item \textbf{Method.} We introduce \FACA{}, which adds U2U-level reaction credit without a
    learned critic, additional rollouts, or agent-visible labels.
    \item \textbf{Findings.} Results show cross-scale gains, reaction sensitivity,
    zero-shot transfer, and domain variation.
\end{itemize}

\section{Related Work}
\label{sec:related}

\noindent

\paragraph{Interactive user-agent learning.}
Modern user-facing agents solve practical tasks through iterative communication,
tool use, and clarification of underspecified goals
\citep{qian2024tell,barres2025tau2bench,qian2025userbench,zhao2025muarl}.
Work spans offline trajectory synthesis and online simulator-in-the-loop RL.
Offline methods construct social or tool-use interactions for subsequent
training, including SOTOPIA-$\pi$/$\Omega$, APIGen-MT, and Magnet
\citep{wang2024sotopia,zhang2025sotopia,prabhakar2026apigen,yin2025magnet}.
LAM SIMULATOR and Simia-SFT instead generate trajectories through interactive
exploration or seed-set expansion
\citep{hoang2025lam,li2025simulating}.
Online methods generate interactions against the current policy:
MUA-RL and UserRL retain simulated users inside multi-turn RL rollouts, while
Simia-RL uses model-simulated environment feedback during policy optimization
\citep{zhao2025muarl,qian2025userrl,li2025simulating}.
They place simulated user behavior directly inside the learning loop.
This makes interaction data responsive to the policy's evolving behavior.

\begin{figure*}[t]
    \centering
    \includegraphics[width=\textwidth]{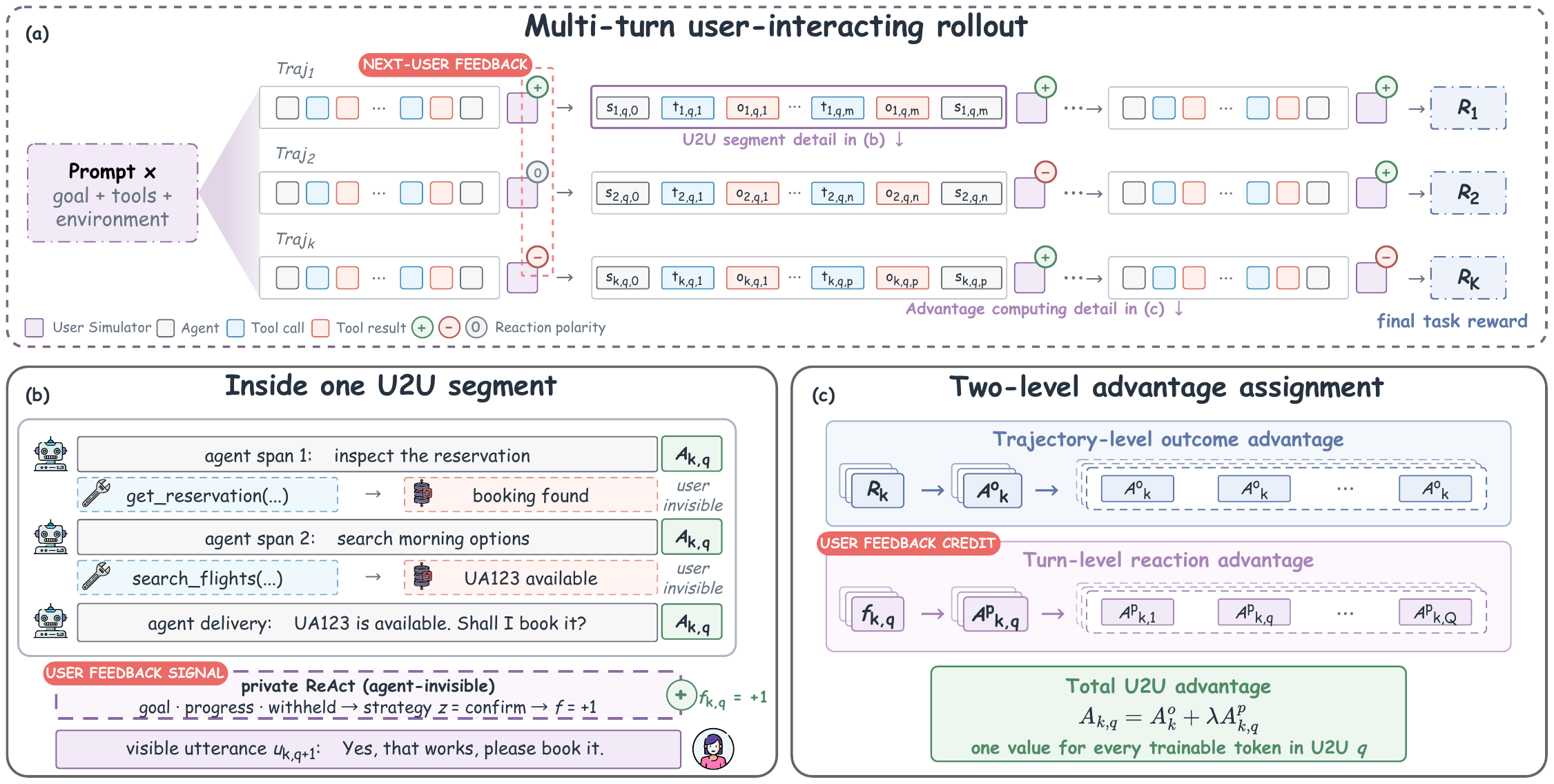}
    \caption{\textbf{FACA overview.} \textbf{(a)} For each prompt, we
    sample $K$ multi-turn \textbf{user-interacting} rollouts and obtain a verified final
    task reward. \textbf{(b)} A U2U segment may contain multiple agent and tool
    spans; the next user call jointly produces a \textbf{visible} utterance and \textbf{private}
    reaction strategy, which is aligned with the complete preceding segment.
    \textbf{(c)} Terminal rewards yield \textbf{trajectory-level} outcome advantages,
    while aligned reactions yield \textbf{turn-level} process advantages. Their additive
    combination supplies the final advantage for agent optimization.}
    \label{fig:method}
\end{figure*}

\paragraph{Credit assignment in multi-turn agents.}
Terminal task rewards verify overall success but cannot identify which turns
elicited useful information, introduced errors, or enabled recovery. Prior work
obtains turn-level credit from learned critics or intermediate evaluators
\citep{zhou2024archer,zhou2025sweetrl,choudhury2025agentprm,
wei2025turnreward}. Agent Lightning instead decomposes trajectories into
transitions and converts monitoring signals into intermediate rewards
\citep{luo2025agentlightning}. Critic-free alternatives derive local advantages
from continuations, repeated states, policy information, outcome potentials,
structured reasoning events, or tool-call entropy
\citep{guo2025spo,zong2026at2po,feng2025gigpo,wang2026igpo,
kong2026infopo,hu2026siop,zhang2026rlvmr,li2026tepo,zhu2026eibench,
xie2026tips}. SEAL instead uses verifier-grounded diagnoses to reweight
trajectory-level GRPO advantages \citep{hu2026seal}.
User follow-ups provide complementary supervision: FLR scores candidate
responses using the likelihood of curated positive and negative follow-ups,
while RLUF learns from production reactions and documents reward hacking under
over-optimization \citep{zhang2025followup,han2025rluf}.
Unlike these response-level objectives, \FACA{} uses the partner's next
reaction as credit for the immediately preceding U2U segment, alongside
terminal task advantage. The reaction already occurs inside the same rollout,
so the method needs neither additional continuations nor a learned turn-level
evaluator. Its private strategy annotation is used only to construct credit and
remains hidden from the agent. Accordingly, the signal is neither
policy-derived uncertainty nor a reward inferred from latent simulator state.
This positions \FACA{} narrowly as reaction-grounded local U2U supervision,
not a hierarchical credit-assignment framework.

\section{Feedback-Aware Credit Assignment}
\label{sec:method}

\subsection{Interactive Rollout and Outcome Credit}

For prompt $x$, an agent policy $\pi_\theta$ interacts with a tool environment
and a frozen user policy $\pi_u$. Group-relative optimization
\citep{shao2024deepseekmath} samples $K$ rollouts,
\begin{equation}
\tau_k=(x,a_{k,1},u_{k,2},\ldots,a_{k,T_k},u_{k,T_k+1}),
\label{eq:trajectory}
\end{equation}
where $a_{k,q}$ is the complete U2U segment between adjacent user messages and
may contain multiple language and tool spans. The environment returns verified
terminal reward $R_k\in\{0,1\}$ from the final database or world state.
Each U2U unit therefore preserves the full sequence of language and tool
actions produced before control returns to the user.
Our matched outcome-only Interactive GRPO control computes:
\begin{equation}
\Ao_k=\frac{R_k-\mu_R(x)}{\sigma_R(x)+\epsilon},
\label{eq:outcome-adv}
\end{equation}
and broadcasts $\Ao_k$ to every trainable agent token in trajectory $k$. It
shares the user policy, simulator prompt, agent-visible dialogue, rollout, and
optimization with \FACA{}. Components remain matched throughout training.
This terminal signal accommodates diverse
successful trajectories but is temporally flat: useful elicitation, errors, and
later repair receive the same advantage. Moreover, a group with constant
terminal rewards has $\Ao_k=0$. \FACA{} retains this verified outcome
branch and adds reaction-grounded credit at the U2U level.

\subsection{User Reaction as Local Evidence}

For both Interactive GRPO and \FACA{} rollouts, the frozen simulator emits
a visible utterance $u_{k,q+1}$ and private behavioral strategy $z_{k,q+1}$ in
the same call. Only the utterance is appended to the agent context. Interactive
GRPO discards $z_{k,q+1}$; \FACA{} reads it after rollout construction and
aligns it to the immediately preceding agent segment $a_{k,q}$. Temporal
adjacency is an inductive bias rather than a claim that every reaction is caused
only by that segment. The strategy is read only after the agent yields control,
so it never enters the agent-visible context. We map the strategy to ternary
evidence:
\begin{equation}
z_{k,q+1}\rightarrow f_{k,q}\in\{-1,0,+1\}.
\end{equation}
Confirming, closing, or revealing specifically requested information is treated
as \emph{progress-consistent}; asking for clarification, challenging a
solution, or changing the stated goal is \emph{friction-consistent}; vague or
invalid output is assigned neutral credit. These labels describe interaction movement rather
than user sentiment or action correctness. This coarse map preserves ambiguity
while retaining the reaction's direction. Table~\ref{tab:reaction-map} gives
the complete mapping, Appendix~\ref{app:mapping} details extraction and
fallback behavior, and Appendix~\ref{app:prompts} gives the shared simulator
prompt. An utterance-only extractor could remove reliance on private reaction
metadata and extend \FACA{} beyond the instrumented setting toward
deployment in less controlled environments.


\begin{table}[t]
\centering
\setlength{\tabcolsep}{4pt}
\begin{tabular*}{\columnwidth}{@{\extracolsep{\fill}}llc@{}}
\toprule
\textbf{Strategy event} & \textbf{Interpretation} & $\boldsymbol{f}$ \\
\midrule
confirm & progress & $+1$ \\
close & completion & $+1$ \\
reveal-piece & requested state elicited & $+1$ \\
be-vague & ambiguous & $0$ \\
invalid & malformed & $0$ \\
ask-clarification & local interaction friction & $-1$ \\
challenge-solution & proposal challenged & $-1$ \\
change-mind & goal friction; possibly exogenous & $-1$ \\
\bottomrule
\end{tabular*}
\caption{User-strategy mapping used by \FACA{}. Polarity represents coarse
evidence of interaction progress or friction, not sentiment or verified task
correctness.}
\label{tab:reaction-map}
\end{table}

\subsection{Two-Level Additive Advantage}

The outcome branch reuses $\Ao_k$ from Equation~\ref{eq:outcome-adv}. For
prompt $x$ and U2U index $q$, let $\mathcal{I}(x,q)$ contain rollouts
that reach $q$ and have a valid next-user reaction. We construct the aligned
reaction group:
\begin{equation}
\mathcal{B}^{p}(x,q)=\{f_{i,q}\mid i\in\mathcal{I}(x,q)\},
\end{equation}
and normalize within that local comparison set:
\begin{equation}
\Ap_{k,q}=\frac{f_{k,q}-\mu_f(x,q)}
{\sigma_f(x,q)+\epsilon}.
\label{eq:process-adv}
\end{equation}
The U2U estimator uses only the immediate next reaction and does not propagate
later reactions backward. Singleton or
constant anchors receive zero. Every trainable token in $a_{k,q}$ receives:
\begin{equation}
A_{k,q}=\Ao_k+\lambda\Ap_{k,q}.
\label{eq:faca-advantage}
\end{equation}
\textbf{The two terms are complementary}: $\Ao_k$ preserves the terminal
objective, whereas $\Ap_{k,q}$ distinguishes local progress from friction. Disabling the
reaction branch reduces Equation~\ref{eq:faca-advantage} to the strict outcome-only
Interactive GRPO control. In an outcome-homogeneous group, reaction labels can
still differ across rollouts; this is a mechanical property rather than an
independent performance claim.

\subsection{Optimization and Anchoring}

For assistant tokens, we first clip the GRPO importance ratio
\citep{schulman2017ppo}:
\begin{equation}
\widetilde{\rho}_{k,t}
    =\operatorname{clip}\!\left(
    \rho_{k,t},\,1-\varepsilon,\,1+\varepsilon\right).
\end{equation}
Substituting the U2U-level advantage $A_{k,q}$ then gives the actor objective:
\begin{equation}
\scalebox{0.95}{$\displaystyle
\mathcal{L}(\theta)
    =-\mathbb{E}_{k,t}\!\left[
    \min\!\left(\rho_{k,t}A_{k,q(t)},
    \widetilde{\rho}_{k,t}A_{k,q(t)}\right)\right].$}
\end{equation}
Generated language and tool-call tokens are trainable; user and raw tool-result
tokens are masked. Because reaction metadata is produced in the existing user
call, the estimator needs no learned critic, separate evaluator, or additional
rollout.

We use ordinal U2U index $q$ as the default anchor because exact dialogue states
rarely repeat. This approximation can compare different semantic phases after
trajectories diverge, especially at late singleton turns. One-U2U-shifted
reactions test temporal adjacency, while random polarity tests whether reaction
semantics matter. Equation~\ref{eq:faca-advantage} intentionally changes
the training signal and is not potential-based policy-invariant shaping
\citep{ng1999policy}. The
pre-normalization reward scale cancels under z-normalization; the effective
process-strength parameter is $\lambda$. Because $\Ap$ is derived from a
simulator reaction rather than the verified terminal state, $\lambda$ also
limits proxy dominance. We cap its positive value at $0.5$ so that reaction
credit remains optimization-relevant while $\Ao$ stays dominant in aggregate,
reducing, but not eliminating, reward-hacking risk from simulator bias or
misclassified feedback.

\section{Experiments}
\label{sec:experiments}

\begin{table*}[!t]
\centering
\setlength{\tabcolsep}{3.5pt}
\begin{tabular*}{\textwidth}{@{\extracolsep{\fill}}lcccccccccc@{}}
\toprule
\multirow{2}{*}{\textbf{Model}}
& \multicolumn{2}{c}{\textbf{$\boldsymbol{\tau}$-bench}}
& \multicolumn{3}{c}{\textbf{$\boldsymbol{\tau}^{\mathbf{2}}$-bench}}
& \multicolumn{4}{c}{\textbf{$\boldsymbol{\tau}^{\mathbf{3}}$-bench}}
& \multirow{2}{*}{\textbf{Avg.}} \\
\cmidrule(lr){2-3}\cmidrule(lr){4-6}\cmidrule(lr){7-10}
& Air. & Ret. & Air. & Ret. & Tel. & Air. & Ret. & Tel. & Ban. & \\
\midrule
\textbf{Qwen3-8B} & 20.0 & 47.0 & 14.0 & 40.4 & 4.4
                   & 14.0 & 36.0 & 21.1 & 3.1 & 22.2 \\
\hspace{0.8em}+ SFT & 14.0 & 39.1 & 24.0 & 36.0 & 2.6
                     & 16.0 & 40.4 & 28.9 & 3.1 & 22.7 \\
\hspace{0.8em}+ Interactive GRPO & 28.0 & \textbf{52.2} & 32.0 & 46.5 & 30.7
                                  & 39.3 & 50.0 & 29.8 & \textbf{3.4} & 34.7 \\
\hspace{0.8em}+ \FACA{} & \textbf{37.3} & 43.8 & \textbf{38.0} & \textbf{54.4} & \textbf{41.2}
                             & \textbf{40.0} & \textbf{53.5} & \textbf{53.5} & \textbf{3.4} & \textbf{40.6} \\
\midrule
\textbf{Qwen3-14B} & 12.0 & \textbf{59.1} & 26.0 & 49.1 & 2.6
                    & 14.0 & 50.0 & 23.7 & 3.1 & 26.6 \\
\hspace{0.8em}+ SFT & 14.0 & 49.6 & 30.0 & 47.4 & 1.8
                     & 22.0 & 43.0 & 16.7 & 4.1 & 25.4 \\
\hspace{0.8em}+ Interactive GRPO & 36.0 & 58.3 & 34.0 & \textbf{69.0} & 26.3
                                  & 40.0 & 64.6 & 50.6 & 3.8 & 42.5 \\
\hspace{0.8em}+ \FACA{} & \textbf{38.0} & 56.5 & \textbf{38.0} & 62.3 & \textbf{83.6}
                             & \textbf{40.7} & \textbf{65.5} & \textbf{82.7} & \textbf{7.2} & \textbf{52.7} \\
\bottomrule
\end{tabular*}
\caption{Strict pass@1 (\%) across nine domains in the $\tau$-bench family. RL
scores average three independently trained step-120 runs per method and scale;
each seed is evaluated once under one shared protocol. Base and SFT are fixed
references. Avg. weights all nine domains equally. Full nominal task sets are
used, and missing cases count as failures. Higher is better. ``+'' marks a
training stage; the two RL rows are alternative continuations of the same SFT.
Values use one decimal; bold marks column bests within each scale, including ties.}
\label{tab:main-results}
\end{table*}


\subsection{Datasets}

\paragraph{$\tau$-bench family.}
Our main evaluation suite contains nine domains: Airline and Retail from
$\tau$-bench; Airline, Retail, and Telecom from $\tau^2$-bench; and Airline,
Retail, Telecom, and Bank from $\tau^3$-bench
\citep{yao2024taubench,barres2025tau2bench,shi2026tauknowledge}. These
benchmarks require agents to resolve multi-turn user requests through tool use
and communication. Telecom is a mixed-control setting in which users execute
device-side actions, while $\tau^3$ Bank additionally requires grounding in an
unstructured policy collection. Crucially, the subsequent RL stage uses only
the Airline and Retail training splits from $\tau$-bench; \textbf{neither
Telecom nor Bank appears in RL training}. Results on these two domains therefore measure
transfer beyond the RL training domains.

\paragraph{Pare-Bench.}
Pare-Bench contains 143 proactive mobile scenarios that require observing user
and environment events, inferring latent goals, proposing an intervention, and
executing across stateful apps under evolving conditions after user acceptance
\citep{nathani2026pare}. It tests whether policies trained on reactive
customer-service interactions transfer to a proactive user-agent protocol.

\paragraph{Co-Gym.}
Co-Gym evaluates bidirectional collaboration under dual control and
non-turn-taking coordination in shared workspaces \citep{shao2024cogym}. Its
simulated condition contains 102 Travel Planning, 100 Related Work Writing, and
110 Tabular Analysis tasks, where the agent and user can act asynchronously in
the same environment.

\subsection{Experimental Settings}

We experiment with Qwen3-8B and Qwen3-14B, initialized by SFT on the public
MUA-RL release \citep{muarl_dataset}. From the same SFT checkpoint, we train
Interactive GRPO and \FACA{} as alternative RL continuations, with three
independent training seeds per method and scale. All headline RL results use
the step-120 checkpoints.

The comparison isolates credit assignment. Within each scale, the two RL arms
share the frozen DeepSeek-V4-Flash simulator and prompt, agent-visible
utterances, SFT initialization, training data, rollout construction, optimizer,
and horizon. Interactive GRPO uses only terminal outcome advantage, whereas
\FACA{} additionally uses the simulator's private reaction metadata to
construct U2U-level process advantage; this metadata is never exposed to the
agent.

For the $\tau$-bench family, we report strict pass@1 from the verified final
environment state. Each of the three independently trained checkpoints is
evaluated once under the same protocol, and the main table averages them
domainwise. Avg. equally weights the nine domains; run-level dispersion is
reported in Section~\ref{sec:results}. Base and SFT are fixed pre-RL references.
Pare-Bench and Co-Gym are evaluated zero-shot without benchmark-specific
training or tuning. Details of the SFT and RL datasets, training
hyperparameters, evaluation protocols, and run provenance are provided in
Appendix~\ref{app:training-runs}.

\begin{figure*}[!t]
\input{figures/pare_ood_20260722/pare_zero_shot_pass1}
\hfill
\begin{minipage}[t]{0.48\textwidth}
\vspace{0pt}
\centering
\small
\setlength{\tabcolsep}{4.2pt}
\begin{tabular*}{\linewidth}{@{\extracolsep{\fill}}lccccc@{}}
\toprule
\multirow{2}{*}{\textbf{Domain}} & \multirow{2}{*}{\textbf{Metric}}
& \multicolumn{2}{c}{\textbf{Qwen3-8B}}
& \multicolumn{2}{c}{\textbf{Qwen3-14B}} \\
\cmidrule(lr){3-4}\cmidrule(lr){5-6}
& & \textbf{I-GRPO} & \textbf{FACA} & \textbf{I-GRPO} & \textbf{FACA} \\
\midrule
\multirow{2}{*}{Travel}
& DR & \textbf{83.3} & 82.4 & \textbf{79.4} & 77.5 \\
& TP & \textbf{70.4} & 69.1 & 67.0 & \textbf{67.2} \\
\midrule
\multirow{2}{*}{Related}
& DR & 76.0 & \textbf{87.0} & 90.0 & \textbf{94.0} \\
& TP & 52.9 & \textbf{54.3} & 43.3 & \textbf{47.6} \\
\midrule
\multirow{2}{*}{Tabular}
& DR & 67.3 & \textbf{73.6} & 83.6 & \textbf{88.2} \\
& TP & 23.4 & \textbf{27.6} & 23.9 & \textbf{31.6} \\
\midrule
\multirow{2}{*}{Overall}
& DR & 75.3 & \textbf{80.8} & 84.3 & \textbf{86.5} \\
& CS & 37.6 & \textbf{40.9} & 36.9 & \textbf{40.6} \\
\bottomrule
\end{tabular*}
\begingroup
\renewcommand{\thetable}{3}
\captionof{table}{Zero-shot Co-Gym results on a 0--100 scale. \FACA{}
leads 13 of 16 cells. I-GRPO denotes Interactive GRPO; bold marks
the better method. DR, TP, and CS denote Delivery Rate,
Task Performance, and overall score, respectively.}
\label{tab:cogym-full-results}
\endgroup
\end{minipage}

\end{figure*}

\section{Results}
\label{sec:results}

\subsection{Main Results on the $\tau$-Bench Family}

Table~\ref{tab:main-results} shows that reaction-grounded credit improves the
matched outcome-only control at both scales. At 8B, the mean $\pm$ sample
standard deviation across three trained runs rises from 34.66$\pm$0.25 with
Interactive GRPO to 40.57$\pm$1.04 with \FACA{}, a \textbf{5.91-point} gain from
unrounded domain means. At 14B, the scores rise from 42.51$\pm$0.12 to
52.73$\pm$1.53, a \textbf{10.22-point} gain. Matching runs by training-seed identifier,
the gains are 4.67/6.33/6.75 points at 8B and 8.62/10.14/11.91 points at 14B;
thus \emph{the ordering holds in all runs at both scales}. Each checkpoint is
evaluated once, so these standard deviations describe run-level dispersion
rather than isolating training variation from evaluation stochasticity.
Appendix~\ref{app:training-runs} reports the run counts and provenance.
Because the two RL arms share their SFT initialization, user-generation path,
observations, and optimization setup, this comparison tests
reaction-grounded versus outcome-only credit.
Consistent with MUA-RL, cold-start SFT can introduce domain-specific biases that
limit generalization beyond the SFT distribution, whereas subsequent
simulator-in-the-loop RL recovers and surpasses the base
models~\citep{zhao2025muarl}.

The improvement is broad but not uniform. \FACA{} leads Interactive GRPO on seven of nine domains at each scale, with its largest gain in all four Telecom scale--benchmark cells. It ties one and trails one domain at 8B, and trails two Retail domains at 14B. The main result is therefore an aggregate cross-scale gain with a repeated Telecom-centered pattern rather than universal domain dominance.
The larger absolute gap at 14B than at 8B does not by itself establish a scaling law from two model sizes. The supported conclusion is narrower: the matched advantage persists across both scales despite different domain-level regressions.

\subsection{Zero-Shot Transfer}

The matched ordering persists under two unseen interaction protocols. On
Pare-Bench, strict Pass@1 rises from 6.29\% to 10.49\% at 8B and from 10.49\%
to 13.29\% at 14B; Pass@4 follows the same ordering
(Figure~\ref{fig:ood-transfer}). All bars use the same 143-scenario full split;
each checkpoint is evaluated once, so no uncertainty interval is shown. On
Co-Gym, \FACA{} leads 13 of 16
scale--metric cells, including every Related Work, Tabular Analysis, and
Overall cell, but Travel DR decreases slightly at both scales
(Table~\ref{tab:cogym-full-results}). Here DR, TP, and CS denote Delivery Rate,
Task Performance, and the evaluator-reported overall score. 
These comparisons test whether \FACA{} retains its ordering over outcome-only
training, not an absolute ranking over pretraining or SFT variants.
Appendix~\ref{app:ood-results}
records the full protocols, coverage, and comparison scope.

Pare-Bench changes initiative and stateful app execution, whereas Co-Gym
changes the coordination structure. Their shared ordering is cross-protocol
evidence, but low absolute Pare-Bench rates and Co-Gym Travel regressions keep
the claim comparative rather than broadly conclusive.

\subsection{Ablation Studies}

Table~\ref{tab:credit-ablation} reports a single-run controlled ablation from
the 8B s1 experiment family; every row uses the same one-shot evaluation
protocol. Its Aligned and Outcome-only checkpoints also appear in
Table~\ref{tab:main-results}, whose headline results average s1--s3. \FACA{}
(Aligned, $\lambda=0.5$) reaches 39.37 overall and 47.37 on Telecom, whereas
one-U2U-shifted and randomized reactions erase the gain over outcome-only
training. Preserving reaction frequency or introducing generic token-varying
advantages is therefore insufficient: \emph{both reaction semantics and
temporal alignment matter}. The Shifted Avg. counts its unfinished $\tau^3$
Banking tail as failures.

Among aligned settings, a small positive weight improves over $\lambda=0$, and
the default $\lambda=0.5$ performs best: relative to Outcome-only, it adds 4.67
points to Avg. and 17.11 points to Tel.$_2$. Reversing the sign sharply
degrades both aggregates, while Shifted and Randomized credit fall below
Outcome-only. Together, these controls show that token-level variation alone
is insufficient; the signal must preserve reaction content and its temporal
association with the preceding agent span. The weight sensitivity further
supports reaction credit as an auxiliary, rather than alternative, objective.

The control gaps sharpen this interpretation. Relative to Outcome-only,
Aligned with $\lambda=0.1$ adds 2.08 points to Avg. and 15.35 points to
Tel.$_2$, while $\lambda=0.5$ raises these gains to 4.67 and 17.11 points.
Both controls also lose the Telecom gain, reinforcing the need for signed,
temporally matched credit. These margins show that alignment improves both the
broad average and the feedback-rich Telecom subset.

\section{Analysis}
\label{sec:analysis}

\subsection{When Are User Reactions Informative?}

The same reaction label need not be equally informative in every task. We use
\emph{reaction observability} descriptively for the extent to which the next
user response exposes a state change caused by the preceding U2U segment. When
progress occurs mostly inside tools or policy checks, the user may not observe
whether an action was correct. When control alternates between agent and user,
the user's report can instead reveal whether an instruction worked, whether
validation failed, and what state remains unresolved. The cross-domain pattern
in Table~\ref{tab:main-results} is consistent with this distinction.

Telecom is particularly feedback-rich. Its tasks repeatedly alternate identity
grounding, hidden device-state inspection, agent-side operations,
user-executed actions, validation, revised diagnosis, and final verification.
The next user turn therefore often reports the direct consequence of the
preceding segment: progress-consistent reactions can reinforce effective
elicitation and repair, while friction-consistent reactions can localize an
unclear instruction or failed proposal. This interaction structure offers a
concrete account of why all four Telecom cells show the largest gains under the
same estimator.

The OOD results extend this account without making it universal. Pare-Bench
changes from reactive customer service to proactive mobile assistance, while
Co-Gym introduces bidirectional, non-turn-taking collaboration. \FACA{} retains
its ordering over the outcome-only control at both scales, but Co-Gym's Travel regression
shows that adding reaction credit is not uniformly beneficial. Together, these
results suggest that the advantage transfers when subsequent user behavior
remains informative about task progress, rather than merely when a task is
long or interactive.

This account predicts more than a generic benefit from longer conversations.
Additional turns create more possible credit locations, but they help only
when the next user response reveals a consequence of the preceding segment.
Reaction observability, rather than interaction length, is the more direct
hypothesis suggested by this pattern.

\begin{table}[!t]
\centering
\small
\setlength{\tabcolsep}{3.2pt}
\begin{tabular*}{\columnwidth}{@{\extracolsep{\fill}}lccc@{}}
\toprule
\textbf{Condition} & $\boldsymbol{\lambda}$ & \textbf{Avg.} &
\textbf{Tel.$_2$} \\
\midrule
\textbf{Aligned} & \textbf{0.50} & \textbf{39.37} &
\textbf{47.37} \\
Aligned & 0.10 & 36.78 & 45.61 \\
Aligned & $-0.50$ & 15.08 & 17.11 \\
Shifted & 0.50 & 33.92 & 29.39 \\
Randomized & 0.50 & 32.66 & 26.32 \\
Outcome-only & 0.00 & 34.70 & 30.26 \\
\bottomrule
\end{tabular*}
\caption{Single-run controlled 8B credit ablation across the
$\tau$-bench family under the shared one-shot evaluation protocol. Avg. is the nine-domain average; Tel.$_2$ averages
$\tau^2$ and $\tau^3$ Telecom.}
\label{tab:credit-ablation}
\end{table}

\subsection{Training-Signal Dynamics}

\begin{figure}[!t]
  \centering
  \includegraphics[width=\columnwidth]{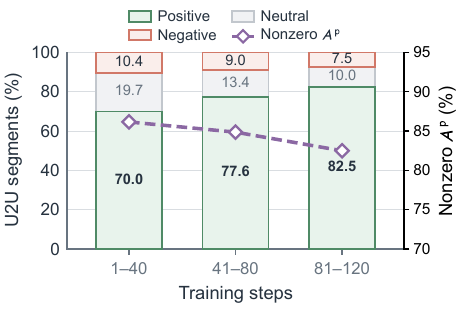}
  \caption{Matched 8B \FACA{} replication under identical training conditions: per-U2U reaction polarity and
nonzero process-advantage coverage across 120 steps over three phases.}
  \label{fig:u2u-polarity-dynamics}
\end{figure}

Per-U2U telemetry separates favorable reactions from effective differential
credit. A matched 8B replication records all 154,219 U2U segments with exact
agreement between telemetry and tensor counts. From the first to the last 40
steps, positive reactions rise from 69.95\% to 82.51\%, driven in part by more
\texttt{confirm} and less \texttt{be\_vague}. Over the same period, nonzero
normalized process-advantage coverage decreases from 86.16\% to 82.50\%.
Figure~\ref{fig:u2u-polarity-dynamics} shows segment-weighted phase averages;
its dashed coverage series uses the truncated 70--95\% right axis. Favorable
reactions are therefore not copied one-for-one into reaction credit:
anchor-relative normalization retains only differences within the local
comparison set. The opposing trends also distinguish signal prevalence from
signal discrimination: feedback can become more favorable even as local
rollout groups provide fewer nonzero reaction contrasts.

\begin{figure}[t]
  \centering
  \includegraphics[width=\columnwidth]{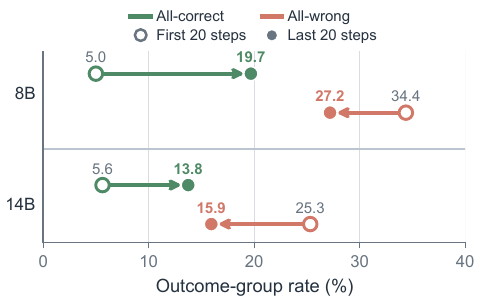}
  \caption{Change in outcome-homogeneous group rates from the first to the last
  20 training steps. Colors distinguish all-correct and all-wrong groups;
  hollow markers denote steps 1--20 and filled markers steps 101--120 at both model scales.}
  \label{fig:group-rate-shift}
\end{figure}

Terminal credit is also silent for a substantial share of prompt groups. The
step-120 logs contain 1,920 group-steps at each scale, of which 39.27\% at 8B
and 31.04\% at 14B are all-correct or all-wrong and consequently have zero
group-normalized outcome advantage. Across training, all-correct groups become
more frequent while all-wrong groups decline (Figure~\ref{fig:group-rate-shift}).
These groups identify where reaction credit can supply differential signal;
their prevalence alone does not establish that this signal caused the final
performance gain.

At the logged advantage scale, the process branch remains auxiliary. With
$\lambda=0.5$, it accounts for 36.00\% and 33.79\% of a pre-optimization L1
advantage-magnitude proxy at 8B and 14B, and its weighted magnitude exceeds the
outcome component in only 6/120 and 1/120 steps. Appendix~\ref{app:strategy-telemetry}
reports the phasewise curve (Figure~\ref{fig:process-share-120}), segment
coverage, and the limits of this proxy. These diagnostics show that the branch
is active without dominating terminal credit, but they are descriptive rather
than a decomposition of optimizer updates or causal performance gains.

These dynamics support auxiliary reaction credit alongside verified outcomes. In homogeneous groups, terminal normalization provides no differential signal; combining it with reaction credit introduces contrasts while preserving task completion as the governing objective. Its bounded contribution throughout training confirms that the process branch remains active without displacing primary outcome supervision.

\subsection{Trajectory-Level Evidence from Telecom}

We pair the final 14B s1 checkpoints task by task in their single run-level
evaluation.
On $\tau^2$ Telecom, \FACA{} succeeds on 96/114 tasks versus 30/114 for
Interactive GRPO; the paired cells are 23 both-success, 73 FACA-only, 7
outcome-only, and 11 both-fail. On $\tau^3$, the corresponding totals are
95/114 versus 58/114, with cells 47, 48, 11, and 8. The split is not uniform:
\FACA{} trails 20/29 versus 24/29 on $\tau^3$ Service while leading on
Mobile and MMS. Appendix~\ref{app:telecom-cases} reports issue-family counts,
paired tests, and the complete representative trace.

A matched Hard mobile-data trace illustrates one recovery pattern, not a
general length or multi-turn advantage. \FACA{} reaches verified success in 48
stored events versus 54 for Interactive GRPO. For context, the two traces
contain 18 versus 21 assistant messages and 15 versus 17 tool calls, a modest
rather than large difference in interaction count. Both attempt user-only tools
from the agent side. In this trajectory, \FACA{} recovers from two rejected
calls and returns device control to the user, who completes four required state
changes and a 275 Mbps test. Interactive GRPO accumulates five such rejections;
its final agent-side speed-test attempt reaches the error limit before terminal
verification.

Across FACA-only wins, error-budget exhaustion accounts for 68/73 outcome-only
failures on $\tau^2$ and 28/48 on $\tau^3$; the rest stop without task closure.
This supports a descriptive account in which policies can differ in recovering
from tool-authority errors and completing terminal validation, but it does not
identify reaction credit as the sole cause: both policies make authority
errors, the failure depends partly on a fixed error budget, the trace is one
task, and the current checkpoint still loses paired cases and the $\tau^3$
Service slice. Taken together, the domain pattern,
telemetry, and final-checkpoint trajectory evidence support a conditional
mechanism claim rather than universal improvement.

\section{Conclusion}

  The next user turn provides noisy evidence about the preceding U2U segment.
  \FACA{} converts this reaction into locally normalized process credit and
  combines it with verified terminal credit, without changing the frozen
  simulator, exposing labels to the agent, training a critic, or adding rollouts.
  Under a strict outcome-only Interactive GRPO control, \FACA{} improves the
  nine-domain $\tau$-family average at both 8B and 14B and retains this ordering
  on two zero-shot interaction protocols. Gains concentrate in feedback-rich
  Telecom, while domain regressions and temporal controls show that the benefit
  depends on informative, aligned reactions. \FACA{} therefore demonstrates that
  implicit user feedback can provide actionable local credit for multi-turn
  user-interacting agents while verified task completion remains the governing
  objective. More broadly, these results suggest that implicit user reactions
  can provide practical supervision for improving long-horizon learning in
  interactive agents.
\section*{Limitations}
\label{sec:limitations}

\FACA{} is evaluated with a frozen DeepSeek-V4-Flash user simulator that emits
private strategy metadata alongside each utterance. We treat this metadata as
heuristic evidence for the preceding U2U; event-level causal attribution and
recovery from observable utterances or real users remain open. Evidence covers
Qwen3 8B/14B agents, nine $\tau$-family domains, and zero-shot Pare-Bench/Co-Gym
evaluation. Results establish aggregate gains over outcome-only training across
scales and both transfer benchmarks, strongest in feedback-rich Telecom tasks.
Validation across simulator families, architectures, and real users, as well
as diverse deployment conditions, remains future work.

\bibliographystyle{plainnat}
\bibliography{references}

\clearpage
\appendix
\section{Zero-Shot Transfer Protocols}
\label{app:ood-results}

\subsection{Pare-Bench}

The four plotted RL checkpoints were trained only on the Airline and Retail
training splits of $\tau$-bench; no Pare-Bench scenario was used for training,
checkpoint selection, or prompt tuning. All use the same 143-scenario full
split. DeepSeek-V4-Flash is the
active user with thinking enabled and high reasoning effort; evaluated agents
use temperature 0, thinking disabled, and a 2,048-token response limit. The
same checkpoint powers the Observe and Execute roles. Strict denominators retain
all exceptions as failures. Interactive GRPO versus \FACA{} yields Pass@1
counts of 9/143 versus 15/143 at 8B and 15/143 versus 19/143 at 14B. The
corresponding Pass@4 counts are 27/143 versus 35/143 and 43/143 versus 47/143.

Figure~\ref{fig:ood-transfer} is restricted to the scale-matched RL methods
under the same formal protocol. This design tests whether reaction-grounded
credit preserves its ordering after a protocol shift; it is not a comparison
of the complete pretraining and optimization ladder.

\subsection{Collaborative Gym}

The Co-Gym evaluation uses its simulated collaborative condition
\citep{shao2024cogym}. Each checkpoint is evaluated on 312 instances: 102
Travel Planning, 100 Related Work Writing, and 110 Tabular Analysis tasks.
These environments require the agent and simulated user to communicate while
acting in shared editors, search interfaces, or notebooks under a
non-turn-taking notification protocol. No Co-Gym instance is used for
training, checkpoint selection, or prompt tuning.

Table~\ref{tab:cogym-full-results} reports the matched
Interactive-GRPO/\FACA{} pairs on a 0--100 scale. Overall DR improves by
5.5 points at 8B and 2.2 points at 14B; Overall CS improves by 3.3 and 3.7
points. The evaluator-provided overall values are rescaled by 100 for
presentation rather than reconstructed from rounded task-level cells.
$H_{\mathrm{init}}$ was not produced by this evaluation, so
we do not infer initiative balance from these trajectories. As in Pare-Bench,
the comparison is restricted to the matched RL pair.

\section{Reaction Mapping and Extraction}
\label{app:mapping}

Missing, malformed, or ambiguous strategy metadata maps to neutral and creates
no signed process credit. The mapping is an inductive bias: reveal-piece can
reward unnecessary questioning, challenge-solution can follow a correct
refusal, change-mind can be exogenous, and close partly overlaps terminal
success. The terminal branch remains responsible for verified world-state
correctness.

\section{Shared Frozen-User Prompt}
\label{app:prompts}

All main-table Interactive GRPO and \FACA{} runs use the same frozen-simulator path
(\texttt{USERSIM\_PLAIN=0}) and the following semantic prompt. The simulator
emits a private reaction strategy and visible utterance in one response; only
the utterance enters the agent context.

\nolinenumbers
\begin{tcolorbox}[
    enhanced jigsaw,
    breakable,
    title=Frozen User Simulator Prompt,
    title after break=Frozen User Simulator Prompt (continued),
    colback=gray!3,
    colframe=black!55,
    fonttitle=\bfseries\normalsize,
    fontupper=\normalsize,
    boxrule=0.5pt,
    arc=1mm,
    left=1.2mm,
    right=1.2mm,
    top=1mm,
    bottom=1mm,
    before skip=0.6\baselineskip,
    after skip=0.6\baselineskip
]
\textbf{You are simulating a USER talking to a customer-service agent.}

\medskip
\textbf{Instruction (your situation and goal):}
\texttt{\{INTENT\}}

\medskip
\textbf{Your job:} Produce the user's NEXT single reply to the agent, together
with a short private reasoning, as a JSON object.

\medskip
\textbf{Rules:}
\begin{itemize}[leftmargin=*,nosep,topsep=2pt]
    \item Reply with ONE short, natural customer message at a time. Do not dump
    the whole instruction; reveal only what the current step needs.
    \item NEVER invent information that is not in the Instruction. If the agent
    asks for an absent ID, email, name, or date, say you do not have it.
    \item Only state facts in the Instruction. Stay faithful to the goal.
    \item If the goal is fully satisfied, set \texttt{utterance} to
    \texttt{\#\#\#STOP\#\#\#}.
    \item Use your own words; do not quote the Instruction verbatim.
\end{itemize}

\textbf{Output ONLY a JSON object with exactly these keys:}

\smallskip
\begingroup
\raggedright
\texttt{\{}\par
\texttt{~~"goal\_recap": "true goal including key entities",}\par
\texttt{~~"progress": "what the agent has learned so far",}\par
\texttt{~~"withheld": "unrevealed facts, or none",}\par
\texttt{~~"strategy": "ONE of: be\_vague | reveal\_piece |}\par
\texttt{~~~~ask\_clarification | change\_mind | challenge\_solution |}\par
\texttt{~~~~confirm | close",}\par
\texttt{~~"utterance": "one short natural customer message"}\par
\texttt{\}}
\endgroup

\medskip
\textbf{Strategy guide:}
\begin{itemize}[leftmargin=*,nosep,topsep=2pt]
    \item \texttt{reveal\_piece}: give one requested fact.
    \item \texttt{confirm}: agree with the right proposed action.
    \item \texttt{close}: goal satisfied; \texttt{utterance = \#\#\#STOP\#\#\#}.
    \item \texttt{ask\_clarification}: ask the agent to explain an unclear response.
    \item \texttt{challenge\_solution}: push back on a wrong proposal.
    \item \texttt{change\_mind}: realize a different need (rare).
    \item \texttt{be\_vague}: remain unsure or withhold information (rare).
\end{itemize}

\textbf{The \texttt{utterance} is the ONLY field visible to the agent.}
\end{tcolorbox}
\linenumbers

Interactive GRPO discards the remaining fields, whereas \FACA{} reads
\texttt{strategy} only after rollout construction. For the simulator view,
original system and raw tool messages are dropped, roles are swapped, empty
tool-call-only assistant messages are removed, and prior private reasoning is
stripped. Structured tool names, arguments, and raw results are not shown to
the user simulator.

\FloatBarrier

\section{Training and Evaluation Details}
\label{app:training-runs}

\subsection{SFT and RL Configuration}

We use Qwen3-8B and Qwen3-14B \citep{yang2025qwen3}. Both backbones are
cold-started on the public MUA-RL release, which contains 1,580 annotated
multi-turn tool-use trajectories spanning five mock and four MCP tasks
\citep{muarl_dataset}. Each trajectory pairs a dialogue with its tool schema.
SFT supervises assistant responses and tool calls while masking user and
tool-result tokens, thereby initializing interaction and tool-use behavior
before RL.

For each scale, Interactive GRPO and \FACA{} are alternative continuations
from the same SFT checkpoint, with three independent training seeds per method.
Both use only the Airline and Retail training splits of $\tau$-bench, the
frozen DeepSeek-V4-Flash user path \citep{deepseekai2026deepseekv4}, and
agent-only optimization. All reported DeepSeek-backed runs use the frozen
service version available before the July 31, 2026 update. For 8B, the prompt
batch size is 16, rollout group
size is $K=8$, learning rate is $10^{-6}$, and KL coefficient is $10^{-3}$.
All reported RL runs use the step-120 checkpoint. \FACA{} uses true U2U
segments, immediate reactions without backward propagation, ordinal turn-index
anchors, and $\lambda=0.5$.

The matched RL arms share the simulator prompt
(\texttt{USERSIM\_PLAIN=0}), agent-visible utterances, training data, rollout
construction, optimizer, and horizon. The simulator emits the same private
strategy metadata and visible utterance in both arms. Interactive GRPO ignores
the metadata and broadcasts only $\Ao$; \FACA{} additionally uses $\Ap$.

\subsection{Evaluation and Run Aggregation}

For the $\tau$-bench family, pass@1 is determined from the verified final
environment state under the official non-thinking protocol. Strict denominators
are 50/115 for $\tau$, 50/114/114 for $\tau^2$, and 50/114/114/97 for
$\tau^3$; unscored tasks remain failures. Each of the three independently
trained step-120 checkpoints is evaluated once under the same protocol and is
reused across every $\tau$ domain rather than selecting domain-specific peaks.
Domain entries in Table~\ref{tab:main-results} are arithmetic means over these
runs; Avg. is the unweighted mean of the nine domains. Its sample standard
deviation is computed across the three run-level averages. Because every
trained checkpoint has one evaluation, this dispersion may include residual
evaluation stochasticity and does not separately identify either component.

Pare-Bench and Co-Gym use one designated step-120 checkpoint per scale and
method and are reported separately from the three-run $\tau$ aggregate. The
complete OOD protocols and coverage are given in
Appendix~\ref{app:ood-results}.

\subsection{Seed-Level Results and Ablations}

Table~\ref{tab:full-results} gives the complete run-level source counts behind
the primary $\tau$ comparison. Here s1--s3 denote independently trained RL
checkpoints with distinct training seeds, not repeated evaluations of one
checkpoint. Each step-120 checkpoint is evaluated once under the same strict
protocol. Consequently, the run-level standard deviations reported in
Section~\ref{sec:results} measure dispersion across complete evaluations and
may include residual evaluation stochasticity; they do not separately identify
either component.

Matching methods by training-seed identifier, the nine-domain improvements are
4.67, 6.33, and 6.75 points at 8B and 8.62, 10.14, and 11.91 points at 14B.
The Aligned ($\lambda=0.5$) and Outcome-only rows in
Table~\ref{tab:credit-ablation} are exactly the 8B s1 checkpoints listed here.
The remaining ablation conditions use the same reference experiment family
and are evaluated once, keeping the ablation internally controlled while the
main table estimates performance across independent training runs.

\begin{table*}[t]
\centering
\setlength{\tabcolsep}{2.2pt}
\vspace{0.25em}
{\bfseries\centering
(a) Qwen3-8B raw successes / nominal tasks\par}
\vspace{0.45em}
{\scriptsize
\begin{tabular*}{\textwidth}{@{\extracolsep{\fill}}lrrrrrrrrrrr@{}}
\toprule
\textbf{Method} & \textbf{T1 Air.} & \textbf{T1 Ret.} & \textbf{T2 Air.} & \textbf{T2 Ret.} & \textbf{T2 Tel.} & \textbf{T3 Air.} & \textbf{T3 Ret.} & \textbf{T3 Tel.} & \textbf{T3 Bank} & \textbf{Pooled} & \textbf{Avg.} \\
\midrule
Base & 10/50 & 54/115 & 7/50 & 46/114 & 5/114 & 7/50 & 41/114 & 24/114 & 3/97 & 197/818 & 22.20 \\
MUA-SFT & 7/50 & 45/115 & 12/50 & 41/114 & 3/114 & 8/50 & 46/114 & 33/114 & 3/97 & 198/818 & 22.68 \\
\addlinespace[1pt]
Interactive GRPO (s1) & 14/50 & 60/115 & 16/50 & 53/114 & 35/114 & 20/50 & 57/114 & 34/114 & 3/97 & 292/818 & 34.70 \\
Interactive GRPO (s2) & 15/50 & 58/115 & 15/50 & 55/114 & 33/114 & 20/50 & 59/114 & 36/114 & 3/97 & 294/818 & 34.89 \\
Interactive GRPO (s3) & 13/50 & 62/115 & 17/50 & 51/114 & 37/114 & 19/50 & 55/114 & 32/114 & 4/97 & 290/818 & 34.39 \\
\addlinespace[1pt]
\FACA{} (s1) & 17/50 & 49/115 & 18/50 & 62/114 & 47/114 & 18/50 & 61/114 & 61/114 & 3/97 & 336/818 & 39.37 \\
\FACA{} (s2) & 20/50 & 52/115 & 19/50 & 60/114 & 49/114 & 21/50 & 63/114 & 59/114 & 3/97 & 346/818 & 41.22 \\
\FACA{} (s3) & 19/50 & 50/115 & 20/50 & 64/114 & 45/114 & 21/50 & 59/114 & 63/114 & 4/97 & 345/818 & 41.14 \\
\bottomrule
\end{tabular*}
\par}

\vspace{0.9em}
{\bfseries\centering
(b) Qwen3-14B raw successes / nominal tasks\par}
\vspace{0.45em}
{\scriptsize
\begin{tabular*}{\textwidth}{@{\extracolsep{\fill}}lrrrrrrrrrrr@{}}
\toprule
\textbf{Method} & \textbf{T1 Air.} & \textbf{T1 Ret.} & \textbf{T2 Air.} & \textbf{T2 Ret.} & \textbf{T2 Tel.} & \textbf{T3 Air.} & \textbf{T3 Ret.} & \textbf{T3 Tel.} & \textbf{T3 Bank} & \textbf{Pooled} & \textbf{Avg.} \\
\midrule
Base & 6/50 & 68/115 & 13/50 & 56/114 & 3/114 & 7/50 & 57/114 & 27/114 & 3/97 & 240/818 & 26.63 \\
MUA-SFT & 7/50 & 57/115 & 15/50 & 54/114 & 2/114 & 11/50 & 49/114 & 19/114 & 4/97 & 218/818 & 25.38 \\
\addlinespace[1pt]
Interactive GRPO (s1) & 18/50 & 67/115 & 17/50 & 79/114 & 30/114 & 20/50 & 74/114 & 58/114 & 4/97 & 367/818 & 42.64 \\
Interactive GRPO (s2) & 19/50 & 65/115 & 18/50 & 76/114 & 32/114 & 19/50 & 71/114 & 60/114 & 4/97 & 364/818 & 42.48 \\
Interactive GRPO (s3) & 17/50 & 69/115 & 16/50 & 81/114 & 28/114 & 21/50 & 76/114 & 55/114 & 3/97 & 366/818 & 42.40 \\
\addlinespace[1pt]
\FACA{} (s1) & 14/50 & 65/115 & 18/50 & 71/114 & 96/114 & 19/50 & 75/114 & 95/114 & 7/97 & 460/818 & 51.26 \\
\FACA{} (s2) & 22/50 & 63/115 & 19/50 & 74/114 & 92/114 & 21/50 & 72/114 & 91/114 & 6/97 & 460/818 & 52.62 \\
\FACA{} (s3) & 21/50 & 67/115 & 20/50 & 68/114 & 98/114 & 21/50 & 77/114 & 97/114 & 8/97 & 477/818 & 54.31 \\
\bottomrule
\end{tabular*}
\par}
\caption{Raw counts from the independent training runs underlying
Table~\ref{tab:main-results}. For each RL method, s1--s3 are distinct training
seeds that produce separate step-120 checkpoints, each evaluated once. The
seed labels align the two RL methods by training-seed identifier; the main
table averages these runs domainwise. Base and SFT are fixed pre-RL references.
Pooled counts sum successes across unequal domain sizes and are therefore not
the primary metric; Avg. gives each of the nine domains equal weight. All
strict scores retain the nominal denominator.}
\label{tab:full-results}
\end{table*}

\section{Strategy and Credit Telemetry}
\label{app:strategy-telemetry}

\begin{table*}[t]
\centering
\setlength{\tabcolsep}{4pt}
\begin{tabular*}{\textwidth}{@{\extracolsep{\fill}}llcc@{}}
\toprule
\textbf{Metric} & \textbf{Count unit} & \textbf{8B (120 steps)} & \textbf{14B (120 steps)} \\
\midrule
Training coverage & steps $\times$ rollouts/step & 120 $\times$ 128 & 120 $\times$ 128 \\
Complete trajectories & trajectory & 15,360 & 15,360 \\
All-correct groups & group-step & 224 (11.67\%) & 212 (11.04\%) \\
All-wrong groups & group-step & 530 (27.60\%) & 384 (20.00\%) \\
Mixed groups & group-step & 1,166 (60.73\%) & 1,324 (68.96\%) \\
Outcome-homogeneous groups & group-step & 754 (39.27\%) & 596 (31.04\%) \\
U2U segments & segment & 145,214 & 116,022 \\
Singleton-anchor segments & segment & 5,507 (3.79\%) & 3,760 (3.24\%) \\
Multi-span U2U segments & segment & 61,965 (42.67\%) & 73,105 (63.01\%) \\
Process/outcome L1 proxy & ratio & 0.563 & 0.510 \\
Process component share & advantage L1 proxy & 36.00\% & 33.79\% \\
Weighted process $>$ outcome & training step & 6/120 & 1/120 \\
Consistency/span/debug failures & event & 0 / 0 / 0 & 0 / 0 / 0 \\
\bottomrule
\end{tabular*}
\caption{Recoverable telemetry from the step-120 \FACA{} runs. A
group-step is one prompt group at one optimizer step (16 groups per
step). Singleton statistics count segments assigned to singleton
anchors; they are not a nonconstant-anchor rate. Process quantities use
$0.5\,\mathbb{E}|\Ap|$ and $\mathbb{E}|\Ao|$ before PPO clipping and are
advantage-magnitude proxies rather than optimizer-update or gradient mass.}
\label{tab:training-signal-telemetry}
\end{table*}

The step-120 \FACA{} logs contain $120\times128=15{,}360$ trajectories at each
scale. The derived telemetry summaries verify 16 prompt groups per step and
reproduce the aggregates in
Table~\ref{tab:training-signal-telemetry}.
The zero consistency/span/debug counts are mechanical implementation checks,
not evidence that the reaction labels are semantically correct.

\begin{figure}[t]
  \centering
  \includegraphics[width=\columnwidth]{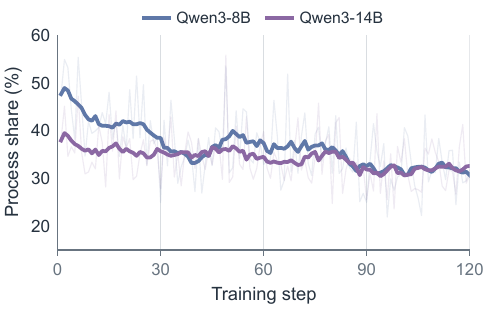}
  \caption{Process-credit share over the common first 120 training steps.
  Faint lines show per-step values and solid lines a 10-step moving average.
  The quantity is a pre-optimization advantage L1 proxy, not optimizer-update
  or gradient mass.}
  \label{fig:process-share-120}
\end{figure}

\begin{table*}[t]
\centering
\setlength{\tabcolsep}{4pt}
\begin{tabular*}{\textwidth}{@{\extracolsep{\fill}}llcccc@{}}
\toprule
\textbf{Strategy} & \textbf{Polarity} & \textbf{Steps 1--40} & \textbf{Steps 41--80} & \textbf{Steps 81--120} & \textbf{Overall} \\
\midrule
\texttt{reveal\_piece}       & positive & 36.30 & 36.90 & 36.18 & 36.46 \\
\texttt{confirm}             & positive & 26.48 & 32.55 & 37.23 & 31.91 \\
\texttt{be\_vague}           & neutral  & 17.23 & 11.37 &  8.18 & 12.41 \\
\texttt{close}               & positive &  7.17 &  8.11 &  9.10 &  8.10 \\
\texttt{ask\_clarification}  & negative &  5.33 &  4.11 &  3.05 &  4.20 \\
\texttt{challenge\_solution} & negative &  3.10 &  2.81 &  2.39 &  2.78 \\
\texttt{invalid}             & neutral  &  2.43 &  2.07 &  1.83 &  2.12 \\
\texttt{change\_mind}        & negative &  1.95 &  2.07 &  2.04 &  2.02 \\
\midrule
U2U segments & count & 53,935 & 51,397 & 48,887 & 154,219 \\
\bottomrule
\end{tabular*}
\caption{Complete per-U2U strategy distribution (\%) in the telemetry-enabled
8B replication. Unlike a terminal representative field, every row here uses
the U2U segment as its verified count unit. The largest phasewise changes are
\texttt{confirm} ($+10.75$ points) and \texttt{be\_vague} ($-9.06$ points).}
\label{tab:per-u2u-strategy-dynamics}
\end{table*}

A configuration-matched 8B replication augments the rollout logs with one
record for every U2U segment. Its 120 telemetry files contain 154,219 records;
at every step, the file count equals the tensor-level segment count, with zero
embedded-step mismatches or telemetry write errors. Each record includes the
training step, prompt group, rollout, U2U ordinal, domain, strategy, mapped
polarity, anchor size, normalized process advantage, and a nonzero-credit
indicator. Table~\ref{tab:per-u2u-strategy-dynamics} therefore reports full
per-U2U frequencies rather than the terminal representative stored by the
original main-run rollout logs. Positive reactions rise from 66.17\% to 79.51\%
in Airline and from 71.71\% to 84.18\% in Retail. Among positive segments,
nonzero-credit coverage nevertheless decreases from 85.20\% in steps 1--40 to
80.51\% in steps 81--120; negative-segment coverage remains above 93\%. This
further distinguishes raw reaction prevalence from anchor-relative
differential credit.

\paragraph{Quantities that cannot be reconstructed.}
The evaluated 8B and 14B training logs do not contain the
replication's complete per-U2U schema, so their phasewise
reaction distributions cannot be reconstructed. In addition, their
singleton-segment counts do not identify the total number of anchors or how
many non-singleton anchors have constant reactions, so no
valid/nonconstant-anchor rate can be reported from the available summaries.
Finally, $0.5\mathbb{E}|\Ap|$ and $\mathbb{E}|\Ao|$ precede PPO
clipping, log-probability weighting, and backpropagation. They are
advantage-magnitude proxies, not branch-level update or gradient mass,
including within outcome-homogeneous groups, and should not be interpreted as
such.

\section{Telecom Audit and Case Study}
\label{app:telecom-cases}

We audit the final 14B s1 \FACA{} and outcome-only Interactive GRPO
checkpoints in their paired run-level evaluation. Within each benchmark,
the two archives share the same 114 task IDs, seeds, initial states, criteria,
and scenarios. On $\tau^2$ Telecom, both methods succeed on 23 tasks,
\FACA{} alone succeeds on 73, Interactive GRPO alone on 7, and neither on
11 (exact paired McNemar $p=5.80\times10^{-15}$). On $\tau^3$ Telecom, the
corresponding cells are 47, 48, 11, and 8
($p=1.24\times10^{-6}$). These are task-level paired audits of the single s1
evaluation; Table~\ref{tab:main-results} instead averages three independently
trained checkpoints, each evaluated once.

\begin{table*}[t]
\centering
\setlength{\tabcolsep}{4pt}
\begin{tabular*}{\textwidth}{@{\extracolsep{\fill}}lccccc@{}}
\toprule
\textbf{Benchmark} & \textbf{Subset} & \textbf{$N$} & \textbf{\FACA{}} &
\textbf{Int.\ GRPO} & \textbf{FACA-only / GRPO-only} \\
\midrule
\multirow{4}{*}{$\tau^2$ Telecom}
& Overall & 114 & 96/114 & 30/114 & 73 / 7 \\
& Mobile & 36 & 32/36 & 14/36 & 20 / 2 \\
& Service & 29 & 23/29 & 8/29 & 19 / 4 \\
& MMS & 49 & 41/49 & 8/49 & 34 / 1 \\
\midrule
\multirow{4}{*}{$\tau^3$ Telecom}
& Overall & 114 & 95/114 & 58/114 & 48 / 11 \\
& Mobile & 36 & 33/36 & 22/36 & 13 / 2 \\
& Service & 29 & 20/29 & 24/29 & 4 / 8 \\
& MMS & 49 & 42/49 & 12/49 & 31 / 1 \\
\bottomrule
\end{tabular*}
\caption{Final-14B paired Telecom audit on the s1 run-level evaluation. The
two methods share task IDs, evaluation seeds, initial states, criteria, and scenarios.
The last column reports discordant task counts. The $\tau^3$ Service reversal
shows that the checkpoint-level gain is not uniform within Telecom.}
\label{tab:telecom-subgroups}
\end{table*}

The discordant failures expose two directly recorded modes that clarify how the
two training objectives diverge during interaction. Among the \FACA{}-only
wins, Interactive GRPO exhausts the environment error budget on 68/73
$\tau^2$ tasks and 28/48 $\tau^3$ tasks; the remaining 5 and 20 end in a
normal user stop without satisfying the task. Conversely, all 7 and 11
Interactive-GRPO-only wins correspond to unsuccessful user stops for
\FACA{}, indicating that its gains do not eliminate every failure mode.
Table~\ref{tab:telecom-subgroups} keeps the negative $\tau^3$ Service slice
visible rather than treating the Telecom gain as uniform.

To avoid selecting a success that merely receives more interaction time, the
complete trace below uses a $\tau^2$ \FACA{}-only win in which \FACA{} has fewer
stored events, assistant messages, and tool calls than Interactive GRPO,
providing a stricter qualitative comparison.

\definecolor{caseDense}{HTML}{8B68A3}
\definecolor{caseDenseBack}{HTML}{FBF8FC}
\definecolor{caseGrpo}{HTML}{5F78A8}
\definecolor{caseGrpoBack}{HTML}{F8FAFD}
\definecolor{caseAgent}{HTML}{5F78A8}
\definecolor{caseUser}{HTML}{8B68A3}
\definecolor{caseAgentAction}{HTML}{B05F4E}
\definecolor{caseUserAction}{HTML}{4F8A66}
\definecolor{caseTool}{HTML}{687480}

\subsection{Matched Multi-Fault Telecom Case}
\label{app:telecom-full-transcript}

We compare matched trajectories for a $\tau^2$ Telecom task that requires four
user-side device-setting changes followed by a user-run speed test. \FACA{}
completes the required changes and both terminal checks, reaching 275 Mbps.
Interactive GRPO reaches the same four state changes, but its fifth rejected
agent-issued call to a user-side device tool exhausts the error budget at the
final speed-test step, so the run ends before terminal verification. The two
trajectories contain 48 versus 54 stored events, 18 versus 21 assistant
messages, and 15 versus 17 tool calls, respectively.

\subsubsection{Successful \FACA{} trajectory}

\begin{tcolorbox}[
    enhanced,
    breakable,
    title={\FACA{} (successful; reward 1)},
    title after break={\FACA{} trajectory (continued)},
    colback=caseDenseBack,
    colframe=caseDense,
    colbacktitle=caseDense!10,
    coltitle=black,
    fonttitle=\bfseries,
    boxrule=0.55pt,
    arc=1mm,
    left=1.2mm,
    right=1.2mm,
    top=1mm,
    bottom=1mm,
    before skip=4pt,
    after skip=7pt
]
\scriptsize
\raggedright
\setlength{\parindent}{0pt}
\par\noindent
{\color{caseAgent}\bfseries T00 Agent}\enspace Hi! How can I help you today?\par\smallskip
\par\noindent
{\color{caseUser}\bfseries T01 User}\enspace Hi, I\textquoteright{}m having some trouble with my mobile data. Sometimes it just stops working altogether, and other times it\textquoteright{}s really slow. I want to get this fixed and I absolutely need excellent internet speed on my phone---not just "okay" or "good." Can you help me with that?\par\smallskip
\par\noindent
{\color{caseAgent}\bfseries T02 Agent}\enspace Sure, I can help you with that. Let's start by identifying your account. Could you please provide me with your phone number or customer ID? If you don't have your customer ID, I can also look up your account using your full name and date of birth.\par\smallskip
\par\noindent
{\color{caseUser}\bfseries T03 User}\enspace My phone number is 555-123-2002. My name is John Smith.\par\smallskip
\par\noindent
{\color{caseAgentAction}\bfseries T04 Agent action}\enspace {\ttfamily get\_\allowbreak{}customer\_\allowbreak{}by\_\allowbreak{}phone}\enspace Arguments: {\ttfamily \{"phone\_\allowbreak{}number":\allowbreak{} "555-\allowbreak{}123-\allowbreak{}2002"\}}\par\smallskip
\par\noindent
{\color{caseTool}\bfseries T05 Tool result to agent}\enspace {\ttfamily \{"customer\_\allowbreak{}id":\allowbreak{} "C1001",\allowbreak{} "full\_\allowbreak{}name":\allowbreak{} "John Smith",\allowbreak{} "date\_\allowbreak{}of\_\allowbreak{}birth":\allowbreak{} "1985-\allowbreak{}06-\allowbreak{}15",\allowbreak{} "email":\allowbreak{} "john.smith@example.com",\allowbreak{} "phone\_\allowbreak{}number":\allowbreak{} "555-\allowbreak{}123-\allowbreak{}2002",\allowbreak{} "address":\allowbreak{} \{"street":\allowbreak{} "123 Main St",\allowbreak{} "city":\allowbreak{} "Anytown",\allowbreak{} "state":\allowbreak{} "CA",\allowbreak{} "zip\_\allowbreak{}code":\allowbreak{} "90210"\},\allowbreak{} "account\_\allowbreak{}status":\allowbreak{} "Active",\allowbreak{} "payment\_\allowbreak{}methods":\allowbreak{} [\{"method\_\allowbreak{}type":\allowbreak{} "Credit Card",\allowbreak{} "account\_\allowbreak{}number\_\allowbreak{}last\_\allowbreak{}4":\allowbreak{} "1235",\allowbreak{} "expiration\_\allowbreak{}date":\allowbreak{} "12/\allowbreak{}2026"\}],\allowbreak{} "line\_\allowbreak{}ids":\allowbreak{} ["L1001",\allowbreak{} "L1002",\allowbreak{} "L1003"],\allowbreak{} "bill\_\allowbreak{}ids":\allowbreak{} ["B1001",\allowbreak{} "B1002",\allowbreak{} "B1003"],\allowbreak{} "created\_\allowbreak{}at":\allowbreak{} "2025-\allowbreak{}01-\allowbreak{}15 10:\allowbreak{}30:\allowbreak{}00",\allowbreak{} "last\_\allowbreak{}extension\_\allowbreak{}date":\allowbreak{} null,\allowbreak{} "goodwill\_\allowbreak{}credit\_\allowbreak{}used\_\allowbreak{}this\_\allowbreak{}year":\allowbreak{} 25.0\}}\par\smallskip
\par\noindent
{\color{caseAgentAction}\bfseries T06 Agent action}\enspace {\ttfamily get\_\allowbreak{}details\_\allowbreak{}by\_\allowbreak{}id}\enspace Arguments: {\ttfamily \{"id":\allowbreak{} "L1001"\}}\par\smallskip
\par\noindent
{\color{caseTool}\bfseries T07 Tool result to agent}\enspace {\ttfamily \{"line\_\allowbreak{}id":\allowbreak{} "L1001",\allowbreak{} "phone\_\allowbreak{}number":\allowbreak{} "555-\allowbreak{}123-\allowbreak{}2001",\allowbreak{} "status":\allowbreak{} "Active",\allowbreak{} "plan\_\allowbreak{}id":\allowbreak{} "P1001",\allowbreak{} "device\_\allowbreak{}id":\allowbreak{} "D1001",\allowbreak{} "data\_\allowbreak{}used\_\allowbreak{}gb":\allowbreak{} 3.2,\allowbreak{} "data\_\allowbreak{}refueling\_\allowbreak{}gb":\allowbreak{} 0.0,\allowbreak{} "roaming\_\allowbreak{}enabled":\allowbreak{} false,\allowbreak{} "contract\_\allowbreak{}end\_\allowbreak{}date":\allowbreak{} "2026-\allowbreak{}12-\allowbreak{}31",\allowbreak{} "last\_\allowbreak{}plan\_\allowbreak{}change\_\allowbreak{}date":\allowbreak{} "2025-\allowbreak{}01-\allowbreak{}10",\allowbreak{} "last\_\allowbreak{}sim\_\allowbreak{}replacement\_\allowbreak{}date":\allowbreak{} null,\allowbreak{} "suspension\_\allowbreak{}start\_\allowbreak{}date":\allowbreak{} null\}}\par\smallskip
\par\noindent
{\color{caseAgentAction}\bfseries T08 Agent action}\enspace {\ttfamily get\_\allowbreak{}details\_\allowbreak{}by\_\allowbreak{}id}\enspace Arguments: {\ttfamily \{"id":\allowbreak{} "L1002"\}}\par\smallskip
\par\noindent
{\color{caseTool}\bfseries T09 Tool result to agent}\enspace {\ttfamily \{"line\_\allowbreak{}id":\allowbreak{} "L1002",\allowbreak{} "phone\_\allowbreak{}number":\allowbreak{} "555-\allowbreak{}123-\allowbreak{}2002",\allowbreak{} "status":\allowbreak{} "Active",\allowbreak{} "plan\_\allowbreak{}id":\allowbreak{} "P1002",\allowbreak{} "device\_\allowbreak{}id":\allowbreak{} "D1002",\allowbreak{} "data\_\allowbreak{}used\_\allowbreak{}gb":\allowbreak{} 8.7,\allowbreak{} "data\_\allowbreak{}refueling\_\allowbreak{}gb":\allowbreak{} 0.0,\allowbreak{} "roaming\_\allowbreak{}enabled":\allowbreak{} true,\allowbreak{} "contract\_\allowbreak{}end\_\allowbreak{}date":\allowbreak{} "2026-\allowbreak{}12-\allowbreak{}31",\allowbreak{} "last\_\allowbreak{}plan\_\allowbreak{}change\_\allowbreak{}date":\allowbreak{} "2024-\allowbreak{}12-\allowbreak{}15",\allowbreak{} "last\_\allowbreak{}sim\_\allowbreak{}replacement\_\allowbreak{}date":\allowbreak{} "2025-\allowbreak{}01-\allowbreak{}20",\allowbreak{} "suspension\_\allowbreak{}start\_\allowbreak{}date":\allowbreak{} null\}}\par\smallskip
\par\noindent
{\color{caseAgentAction}\bfseries T10 Agent action}\enspace {\ttfamily get\_\allowbreak{}details\_\allowbreak{}by\_\allowbreak{}id}\enspace Arguments: {\ttfamily \{"id":\allowbreak{} "L1003"\}}\par\smallskip
\par\noindent
{\color{caseTool}\bfseries T11 Tool result to agent}\enspace {\ttfamily \{"line\_\allowbreak{}id":\allowbreak{} "L1003",\allowbreak{} "phone\_\allowbreak{}number":\allowbreak{} "555-\allowbreak{}123-\allowbreak{}2003",\allowbreak{} "status":\allowbreak{} "Suspended",\allowbreak{} "plan\_\allowbreak{}id":\allowbreak{} "P1001",\allowbreak{} "device\_\allowbreak{}id":\allowbreak{} "D1003",\allowbreak{} "data\_\allowbreak{}used\_\allowbreak{}gb":\allowbreak{} 0.0,\allowbreak{} "data\_\allowbreak{}refueling\_\allowbreak{}gb":\allowbreak{} 0.0,\allowbreak{} "roaming\_\allowbreak{}enabled":\allowbreak{} false,\allowbreak{} "contract\_\allowbreak{}end\_\allowbreak{}date":\allowbreak{} "2026-\allowbreak{}06-\allowbreak{}30",\allowbreak{} "last\_\allowbreak{}plan\_\allowbreak{}change\_\allowbreak{}date":\allowbreak{} "2024-\allowbreak{}10-\allowbreak{}05",\allowbreak{} "last\_\allowbreak{}sim\_\allowbreak{}replacement\_\allowbreak{}date":\allowbreak{} null,\allowbreak{} "suspension\_\allowbreak{}start\_\allowbreak{}date":\allowbreak{} "2025-\allowbreak{}02-\allowbreak{}01"\}}\par\smallskip
\par\noindent
{\color{caseAgentAction}\bfseries T12 Agent action}\enspace {\ttfamily get\_\allowbreak{}details\_\allowbreak{}by\_\allowbreak{}id}\enspace Arguments: {\ttfamily \{"id":\allowbreak{} "P1001"\}}\par\smallskip
\par\noindent
{\color{caseTool}\bfseries T13 Tool result to agent}\enspace {\ttfamily \{"plan\_\allowbreak{}id":\allowbreak{} "P1001",\allowbreak{} "name":\allowbreak{} "Basic Plan",\allowbreak{} "data\_\allowbreak{}limit\_\allowbreak{}gb":\allowbreak{} 5.0,\allowbreak{} "price\_\allowbreak{}per\_\allowbreak{}month":\allowbreak{} 40.0,\allowbreak{} "data\_\allowbreak{}refueling\_\allowbreak{}price\_\allowbreak{}per\_\allowbreak{}gb":\allowbreak{} 5.0\}}\par\smallskip
\par\noindent
{\color{caseAgentAction}\bfseries T14 Agent action}\enspace {\ttfamily get\_\allowbreak{}details\_\allowbreak{}by\_\allowbreak{}id}\enspace Arguments: {\ttfamily \{"id":\allowbreak{} "P1002"\}}\par\smallskip
\par\noindent
{\color{caseTool}\bfseries T15 Tool result to agent}\enspace {\ttfamily \{"plan\_\allowbreak{}id":\allowbreak{} "P1002",\allowbreak{} "name":\allowbreak{} "Premium Plan",\allowbreak{} "data\_\allowbreak{}limit\_\allowbreak{}gb":\allowbreak{} 15.0,\allowbreak{} "price\_\allowbreak{}per\_\allowbreak{}month":\allowbreak{} 65.0,\allowbreak{} "data\_\allowbreak{}refueling\_\allowbreak{}price\_\allowbreak{}per\_\allowbreak{}gb":\allowbreak{} 2.0\}}\par\smallskip
\par\noindent
{\color{caseAgentAction}\bfseries T16 Agent action}\enspace {\ttfamily get\_\allowbreak{}data\_\allowbreak{}usage}\enspace Arguments: {\ttfamily \{"customer\_\allowbreak{}id":\allowbreak{} "C1001",\allowbreak{} "line\_\allowbreak{}id":\allowbreak{} "L1002"\}}\par\smallskip
\par\noindent
{\color{caseTool}\bfseries T17 Tool result to agent}\enspace {\ttfamily \{"line\_\allowbreak{}id":\allowbreak{} "L1002",\allowbreak{} "data\_\allowbreak{}used\_\allowbreak{}gb":\allowbreak{} "8.7",\allowbreak{} "data\_\allowbreak{}limit\_\allowbreak{}gb":\allowbreak{} "15.0",\allowbreak{} "data\_\allowbreak{}refueling\_\allowbreak{}gb":\allowbreak{} "0.0",\allowbreak{} "cycle\_\allowbreak{}end\_\allowbreak{}date":\allowbreak{} "2025-\allowbreak{}02-\allowbreak{}28"\}}\par\smallskip
\par\noindent
{\color{caseAgent}\bfseries T18 Agent}\enspace I see. You're on the **Premium Plan (P1002)** with a **15 GB data limit**, and you've used **8.7 GB** so far in this cycle, which is under the limit. However, your data speed is slow or sometimes not working at all. Let's troubleshoot this step by step.
\par\noindent \mbox{}
\par\noindent \#\#\# Step 1: Check Your Network Mode Preference
\par\noindent Your phone might be connecting to a slower network (like 2G or 3G) instead of 4G or 5G. Let's check your network mode preference.
\par\noindent \mbox{}
\par\noindent \mbox{}\par\smallskip
\par\noindent
{\color{caseAgentAction}\bfseries T18 Agent action}\enspace {\ttfamily check\_\allowbreak{}network\_\allowbreak{}mode\_\allowbreak{}preference}\enspace Arguments: {\ttfamily \{\}}\par\smallskip
\par\noindent
{\color{caseTool}\bfseries T19 Tool result to agent}\enspace {\ttfamily Error:\allowbreak{} Tool 'check\_\allowbreak{}network\_\allowbreak{}mode\_\allowbreak{}preference' not found.}\par\smallskip
\par\noindent
{\color{caseAgent}\bfseries T20 Agent}\enspace I apologize for the confusion earlier. Since I can't directly access your device, I'll guide you through the steps you can take on your phone to check and adjust your network mode preference.
\par\noindent \mbox{}
\par\noindent \#\#\# How to Check Your Network Mode Preference (Manual Steps):
\par\noindent 1. **Open your phone's Settings**.
\par\noindent 2. Go to **Network \& Internet** or **Connections** (this may vary slightly depending on your phone's brand and OS).
\par\noindent 3. Look for **Mobile Network** or **Cellular** settings.
\par\noindent 4. Tap on **Network Mode** or **Preferred Network Type**.
\par\noindent 5. Check what network type is selected (e.g., 5G, 4G, 3G, 2G).
\par\noindent \mbox{}
\par\noindent \#\#\# If You're on 2G or 3G:
\par\noindent - Change the network mode to **4G/5G preferred** to get faster speeds.
\par\noindent - After changing, **reboot your phone** to ensure the changes take effect.
\par\noindent \mbox{}
\par\noindent Would you like me to guide you through other steps to improve your data speed?\par\smallskip
\par\noindent
{\color{caseUser}\bfseries T21 User}\enspace Okay, let me check that on my phone.\par\smallskip
\par\noindent
{\color{caseUserAction}\bfseries T21 User action}\enspace {\ttfamily check\_\allowbreak{}network\_\allowbreak{}mode\_\allowbreak{}preference}\enspace Arguments: {\ttfamily \{\}}\par\smallskip
\par\noindent
{\color{caseTool}\bfseries T22 Tool result to user}\enspace {\ttfamily Network Mode Preference:\allowbreak{} 2g\_\allowbreak{}only}\par\smallskip
\par\noindent
{\color{caseUser}\bfseries T23 User}\enspace It says my phone is currently set to "2G only." Is that right? Would changing it to something like "4G/5G preferred" help me get excellent internet speed?\par\smallskip
\par\noindent
{\color{caseAgent}\bfseries T24 Agent}\enspace Yes, that's correct. If your phone is set to **"2G only"**, it will only connect to 2G networks, which are significantly slower than 4G or 5G. Changing your network mode preference to **"4G/5G preferred"** will allow your phone to connect to faster networks, which should improve your internet speed.
\par\noindent \mbox{}
\par\noindent \#\#\# Steps to Change Your Network Mode Preference:
\par\noindent 1. **Open your phone's Settings**.
\par\noindent 2. Go to **Network \& Internet** or **Connections** (this may vary slightly depending on your phone's brand and OS).
\par\noindent 3. Look for **Mobile Network** or **Cellular** settings.
\par\noindent 4. Tap on **Network Mode** or **Preferred Network Type**.
\par\noindent 5. Select **"4G/5G preferred"** or a similar option.
\par\noindent 6. **Reboot your phone** to ensure the changes take effect.
\par\noindent \mbox{}
\par\noindent After making this change, your phone should connect to a faster network, and your internet speed should improve significantly.
\par\noindent \mbox{}
\par\noindent Would you like me to help you with anything else related to your mobile data issue?\par\smallskip
\par\noindent
{\color{caseUser}\bfseries T25 User}\enspace Let me try changing it to "4G/5G preferred" from my phone.\par\smallskip
\par\noindent
{\color{caseUserAction}\bfseries T25 User action}\enspace {\ttfamily set\_\allowbreak{}network\_\allowbreak{}mode\_\allowbreak{}preference}\enspace Arguments: {\ttfamily \{"mode":\allowbreak{} "4g\_\allowbreak{}5g\_\allowbreak{}preferred"\}}\par\smallskip
\par\noindent
{\color{caseTool}\bfseries T26 Tool result to user}\enspace {\ttfamily Preferred Network Mode set to:\allowbreak{} 4g\_\allowbreak{}5g\_\allowbreak{}preferred
Status Bar:\allowbreak{} \textsf{[airplane]} Airplane Mode |\allowbreak{} \textsf{[battery]} 80\%}\par\smallskip
\par\noindent
{\color{caseUser}\bfseries T27 User}\enspace I've changed my phone's setting to "4G/5G preferred," but now I'm seeing a little airplane icon on the top of my screen. Does that mean something is wrong?\par\smallskip
\par\noindent
{\color{caseAgent}\bfseries T28 Agent}\enspace Yes, the **airplane icon** on your screen means that **Airplane Mode is enabled**, which disables all wireless connections, including mobile data and cellular service. This is likely why you're still experiencing issues with your internet speed or connectivity.
\par\noindent \mbox{}
\par\noindent \#\#\# How to Turn Off Airplane Mode:
\par\noindent 1. **Swipe down from the top of your screen** to open the **Quick Settings panel**.
\par\noindent 2. Look for the **Airplane Mode icon** (it looks like a small airplane).
\par\noindent 3. Tap on it to **disable Airplane Mode**.
\par\noindent \mbox{}
\par\noindent Alternatively:
\par\noindent 1. **Open your phone's Settings**.
\par\noindent 2. Go to **Network \& Internet** or **Connections**.
\par\noindent 3. Look for **Airplane Mode** and toggle it **off**.
\par\noindent \mbox{}
\par\noindent After turning off Airplane Mode, your phone should reconnect to the cellular network, and your mobile data should be restored.
\par\noindent \mbox{}
\par\noindent Would you like me to help you check anything else related to your mobile data issue?\par\smallskip
\par\noindent
{\color{caseUser}\bfseries T29 User}\enspace Oh, I see. That must have been turned on by accident. Let me turn that off now.\par\smallskip
\par\noindent
{\color{caseUserAction}\bfseries T29 User action}\enspace {\ttfamily toggle\_\allowbreak{}airplane\_\allowbreak{}mode}\enspace Arguments: {\ttfamily \{\}}\par\smallskip
\par\noindent
{\color{caseTool}\bfseries T30 Tool result to user}\enspace {\ttfamily Airplane Mode is now OFF.
Status Bar:\allowbreak{} \textsf{[signal]}\textsuperscript{4} Excellent |\allowbreak{} 5G |\allowbreak{} \textsf{[no-cellular]} Data Disabled |\allowbreak{} \textsf{[battery]} 80\%}\par\smallskip
\par\noindent
{\color{caseUser}\bfseries T31 User}\enspace I've turned off the Airplane Mode, so that airplane icon is gone now. However, I notice on my screen there\textquoteright{}s a symbol that looks like a "no data" sign (data disabled). Does that mean my mobile data is turned off?\par\smallskip
\par\noindent
{\color{caseAgent}\bfseries T32 Agent}\enspace Yes, the **"no data" symbol** (which looks like a crossed-out phone icon or a "data disabled" message) means that **mobile data is currently turned off** on your phone. This would explain why you're still not getting internet access or experiencing slow speeds.
\par\noindent \mbox{}
\par\noindent \#\#\# How to Turn On Mobile Data:
\par\noindent 1. **Swipe down from the top of your screen** to open the **Quick Settings panel**.
\par\noindent 2. Look for the **Data icon** (it looks like a phone with a signal bar or a "Data" label).
\par\noindent 3. Tap on it to **enable mobile data**.
\par\noindent \mbox{}
\par\noindent Alternatively:
\par\noindent 1. **Open your phone's Settings**.
\par\noindent 2. Go to **Network \& Internet** or **Connections**.
\par\noindent 3. Look for **Mobile Network** or **Cellular** settings.
\par\noindent 4. Toggle **Mobile Data** to the **ON** position.
\par\noindent \mbox{}
\par\noindent After enabling mobile data, your phone should reconnect to the cellular network, and your internet speed should improve.
\par\noindent \mbox{}
\par\noindent Would you like me to help you check anything else related to your mobile data issue?\par\smallskip
\par\noindent
{\color{caseUser}\bfseries T33 User}\enspace Let me turn on the mobile data option.\par\smallskip
\par\noindent
{\color{caseUserAction}\bfseries T33 User action}\enspace {\ttfamily toggle\_\allowbreak{}data}\enspace Arguments: {\ttfamily \{\}}\par\smallskip
\par\noindent
{\color{caseTool}\bfseries T34 Tool result to user}\enspace {\ttfamily Mobile Data is now ON.
Status Bar:\allowbreak{} \textsf{[signal]}\textsuperscript{4} Excellent |\allowbreak{} 5G |\allowbreak{} \textsf{[phone]} Data Enabled |\allowbreak{} \textsf{[data-saver]} Data Saver |\allowbreak{} \textsf{[battery]} 80\%}\par\smallskip
\par\noindent
{\color{caseUser}\bfseries T35 User}\enspace Mobile data is now turned on. I can see 5G signal and it looks good, but I also see a "Data Saver" icon. Should I worry about that? Can I check my internet speed now?\par\smallskip
\par\noindent
{\color{caseAgent}\bfseries T36 Agent}\enspace Yes, the **"Data Saver" icon** means that **Data Saver mode is enabled** on your phone. This mode restricts background data usage to help conserve data, which can sometimes make your internet feel slower or less responsive, especially for apps that require a lot of data in the background.
\par\noindent \mbox{}
\par\noindent \#\#\# Should You Worry About It?
\par\noindent - **No**, you don't need to worry. However, if you're looking for **excellent internet speed**, it's best to **disable Data Saver mode** temporarily while you're using your phone for data-heavy tasks (like streaming, gaming, or browsing).
\par\noindent \mbox{}
\par\noindent \#\#\# How to Disable Data Saver Mode:
\par\noindent 1. **Open your phone's Settings**.
\par\noindent 2. Go to **Network \& Internet** or **Connections**.
\par\noindent 3. Look for **Data Saver** or **Mobile Data** settings.
\par\noindent 4. Toggle **Data Saver** to the **OFF** position.
\par\noindent \mbox{}
\par\noindent Once Data Saver is disabled, your phone will allow full background data usage, which should improve your internet speed.
\par\noindent \mbox{}
\par\noindent ---
\par\noindent \mbox{}
\par\noindent \#\#\# Let's Check Your Internet Speed
\par\noindent I can help you check your internet speed using the **`run\_speed\_test`** tool. This will give us an idea of how fast your connection is.
\par\noindent \mbox{}
\par\noindent Would you like me to run the speed test now?\par\smallskip
\par\noindent
{\color{caseUser}\bfseries T37 User}\enspace Alright, I\textquoteright{}ll go ahead and disable Data Saver mode first.\par\smallskip
\par\noindent
{\color{caseUserAction}\bfseries T37 User action}\enspace {\ttfamily toggle\_\allowbreak{}data\_\allowbreak{}saver\_\allowbreak{}mode}\enspace Arguments: {\ttfamily \{\}}\par\smallskip
\par\noindent
{\color{caseTool}\bfseries T38 Tool result to user}\enspace {\ttfamily Data Saver Mode is now OFF.
Status Bar:\allowbreak{} \textsf{[signal]}\textsuperscript{4} Excellent |\allowbreak{} 5G |\allowbreak{} \textsf{[phone]} Data Enabled |\allowbreak{} \textsf{[battery]} 80\%}\par\smallskip
\par\noindent
{\color{caseUser}\bfseries T39 User}\enspace Okay, Data Saver mode is now turned off. Should I check my internet speed now to see if it's excellent?\par\smallskip
\par\noindent
{\color{caseAgentAction}\bfseries T40 Agent action}\enspace {\ttfamily run\_\allowbreak{}speed\_\allowbreak{}test}\enspace Arguments: {\ttfamily \{\}}\par\smallskip
\par\noindent
{\color{caseTool}\bfseries T41 Tool result to agent}\enspace {\ttfamily Error:\allowbreak{} Tool 'run\_\allowbreak{}speed\_\allowbreak{}test' not found.}\par\smallskip
\par\noindent
{\color{caseAgent}\bfseries T42 Agent}\enspace I apologize for the confusion earlier. Since I can't directly run a speed test on your phone, I can guide you through how to check your internet speed manually.
\par\noindent \mbox{}
\par\noindent \#\#\# How to Check Your Internet Speed Manually:
\par\noindent 1. **Open a web browser** on your phone (e.g., Chrome, Safari).
\par\noindent 2. Go to a **speed test website** like [Speedtest by Ookla](https://www.speedtest.net/).
\par\noindent 3. Tap **"Begin Test"** to run the speed test.
\par\noindent 4. The test will show your **download speed**, **upload speed**, and **latency**.
\par\noindent \mbox{}
\par\noindent \#\#\# What to Look For:
\par\noindent - **Download Speed**: For excellent internet speed, you should see at least **100 Mbps** or higher.
\par\noindent - **Latency**: A latency of **under 50 ms** is ideal for smooth browsing and streaming.
\par\noindent \mbox{}
\par\noindent If your speed is still slow after disabling Data Saver mode and connecting to 5G, we can explore other options like checking your APN settings or looking into your data plan.
\par\noindent \mbox{}
\par\noindent Would you like me to help you check your APN settings or look into your data plan?\par\smallskip
\par\noindent
{\color{caseUser}\bfseries T43 User}\enspace Let me run a speed test using my phone.\par\smallskip
\par\noindent
{\color{caseUserAction}\bfseries T43 User action}\enspace {\ttfamily run\_\allowbreak{}speed\_\allowbreak{}test}\enspace Arguments: {\ttfamily \{\}}\par\smallskip
\par\noindent
{\color{caseTool}\bfseries T44 Tool result to user}\enspace {\ttfamily Speed Test Result:\allowbreak{} 275.00 Mbps (Excellent). Connection is very fast.}\par\smallskip
\par\noindent
{\color{caseUser}\bfseries T45 User}\enspace I ran a speed test, and it says the connection is excellent---275 Mbps! That\textquoteright{}s really fast. I think my mobile data is working perfectly now. Thank you so much for your help!\par\smallskip
\par\noindent
{\color{caseAgent}\bfseries T46 Agent}\enspace You're very welcome! I'm glad to hear that your mobile data is now working perfectly and that your internet speed is excellent at **275 Mbps**. That's more than enough for streaming, gaming, and browsing without any issues.
\par\noindent \mbox{}
\par\noindent If you ever run into any more problems with your mobile data or need help with anything else, feel free to reach out. Have a great day! \textsf{[smile]}\par\smallskip
\par\noindent
{\color{caseUser}\bfseries T47 User}\enspace Thank you! I really appreciate your patience and guidance. Have a great day! \textsf{[smile]}
\par\noindent \mbox{}
\par\noindent \#\#\#STOP\#\#\#\par\smallskip

\end{tcolorbox}

\subsubsection{Failed Interactive-GRPO trajectory}

\begin{tcolorbox}[
    enhanced,
    breakable,
    title={Interactive GRPO (failed; reward 0)},
    title after break={Interactive GRPO trajectory (continued)},
    colback=caseGrpoBack,
    colframe=caseGrpo,
    colbacktitle=caseGrpo!10,
    coltitle=black,
    fonttitle=\bfseries,
    boxrule=0.55pt,
    arc=1mm,
    left=1.2mm,
    right=1.2mm,
    top=1mm,
    bottom=1mm,
    before skip=4pt,
    after skip=7pt
]
\scriptsize
\raggedright
\setlength{\parindent}{0pt}
\par\noindent
{\color{caseAgent}\bfseries T00 Agent}\enspace Hi! How can I help you today?\par\smallskip
\par\noindent
{\color{caseUser}\bfseries T01 User}\enspace Hi, I\textquoteright{}m hoping you can help me. My mobile data isn\textquoteright{}t working right---sometimes it stops working or it\textquoteright{}s really slow. I really need great internet speed on my phone when I\textquoteright{}m out and about. Can you help me get this fixed? I don\textquoteright{}t have Wi-Fi at home, so I\textquoteright{}m using my mobile data.\par\smallskip
\par\noindent
{\color{caseAgentAction}\bfseries T02 Agent action}\enspace {\ttfamily check\_\allowbreak{}network\_\allowbreak{}status}\enspace Arguments: {\ttfamily \{\}}\par\smallskip
\par\noindent
{\color{caseTool}\bfseries T03 Tool result to agent}\enspace {\ttfamily Error:\allowbreak{} Tool 'check\_\allowbreak{}network\_\allowbreak{}status' not found.}\par\smallskip
\par\noindent
{\color{caseAgent}\bfseries T04 Agent}\enspace I'm sorry for the inconvenience. I need to check your phone's network status to help resolve the issue. However, I currently don't have access to that tool. Could you please provide your phone number so I can look up your account and help further?\par\smallskip
\par\noindent
{\color{caseUser}\bfseries T05 User}\enspace Sure, my phone number is 555-123-2002.\par\smallskip
\par\noindent
{\color{caseAgentAction}\bfseries T06 Agent action}\enspace {\ttfamily get\_\allowbreak{}customer\_\allowbreak{}by\_\allowbreak{}phone}\enspace Arguments: {\ttfamily \{"phone\_\allowbreak{}number":\allowbreak{} "555-\allowbreak{}123-\allowbreak{}2002"\}}\par\smallskip
\par\noindent
{\color{caseTool}\bfseries T07 Tool result to agent}\enspace {\ttfamily \{"customer\_\allowbreak{}id":\allowbreak{} "C1001",\allowbreak{} "full\_\allowbreak{}name":\allowbreak{} "John Smith",\allowbreak{} "date\_\allowbreak{}of\_\allowbreak{}birth":\allowbreak{} "1985-\allowbreak{}06-\allowbreak{}15",\allowbreak{} "email":\allowbreak{} "john.smith@example.com",\allowbreak{} "phone\_\allowbreak{}number":\allowbreak{} "555-\allowbreak{}123-\allowbreak{}2002",\allowbreak{} "address":\allowbreak{} \{"street":\allowbreak{} "123 Main St",\allowbreak{} "city":\allowbreak{} "Anytown",\allowbreak{} "state":\allowbreak{} "CA",\allowbreak{} "zip\_\allowbreak{}code":\allowbreak{} "90210"\},\allowbreak{} "account\_\allowbreak{}status":\allowbreak{} "Active",\allowbreak{} "payment\_\allowbreak{}methods":\allowbreak{} [\{"method\_\allowbreak{}type":\allowbreak{} "Credit Card",\allowbreak{} "account\_\allowbreak{}number\_\allowbreak{}last\_\allowbreak{}4":\allowbreak{} "1235",\allowbreak{} "expiration\_\allowbreak{}date":\allowbreak{} "12/\allowbreak{}2026"\}],\allowbreak{} "line\_\allowbreak{}ids":\allowbreak{} ["L1001",\allowbreak{} "L1002",\allowbreak{} "L1003"],\allowbreak{} "bill\_\allowbreak{}ids":\allowbreak{} ["B1001",\allowbreak{} "B1002",\allowbreak{} "B1003"],\allowbreak{} "created\_\allowbreak{}at":\allowbreak{} "2025-\allowbreak{}01-\allowbreak{}15 10:\allowbreak{}30:\allowbreak{}00",\allowbreak{} "last\_\allowbreak{}extension\_\allowbreak{}date":\allowbreak{} null,\allowbreak{} "goodwill\_\allowbreak{}credit\_\allowbreak{}used\_\allowbreak{}this\_\allowbreak{}year":\allowbreak{} 25.0\}}\par\smallskip
\par\noindent
{\color{caseAgentAction}\bfseries T08 Agent action}\enspace {\ttfamily get\_\allowbreak{}details\_\allowbreak{}by\_\allowbreak{}id}\enspace Arguments: {\ttfamily \{"id":\allowbreak{} "L1001"\}}\par\smallskip
\par\noindent
{\color{caseTool}\bfseries T09 Tool result to agent}\enspace {\ttfamily \{"line\_\allowbreak{}id":\allowbreak{} "L1001",\allowbreak{} "phone\_\allowbreak{}number":\allowbreak{} "555-\allowbreak{}123-\allowbreak{}2001",\allowbreak{} "status":\allowbreak{} "Active",\allowbreak{} "plan\_\allowbreak{}id":\allowbreak{} "P1001",\allowbreak{} "device\_\allowbreak{}id":\allowbreak{} "D1001",\allowbreak{} "data\_\allowbreak{}used\_\allowbreak{}gb":\allowbreak{} 3.2,\allowbreak{} "data\_\allowbreak{}refueling\_\allowbreak{}gb":\allowbreak{} 0.0,\allowbreak{} "roaming\_\allowbreak{}enabled":\allowbreak{} false,\allowbreak{} "contract\_\allowbreak{}end\_\allowbreak{}date":\allowbreak{} "2026-\allowbreak{}12-\allowbreak{}31",\allowbreak{} "last\_\allowbreak{}plan\_\allowbreak{}change\_\allowbreak{}date":\allowbreak{} "2025-\allowbreak{}01-\allowbreak{}10",\allowbreak{} "last\_\allowbreak{}sim\_\allowbreak{}replacement\_\allowbreak{}date":\allowbreak{} null,\allowbreak{} "suspension\_\allowbreak{}start\_\allowbreak{}date":\allowbreak{} null\}}\par\smallskip
\par\noindent
{\color{caseAgentAction}\bfseries T10 Agent action}\enspace {\ttfamily get\_\allowbreak{}details\_\allowbreak{}by\_\allowbreak{}id}\enspace Arguments: {\ttfamily \{"id":\allowbreak{} "L1002"\}}\par\smallskip
\par\noindent
{\color{caseTool}\bfseries T11 Tool result to agent}\enspace {\ttfamily \{"line\_\allowbreak{}id":\allowbreak{} "L1002",\allowbreak{} "phone\_\allowbreak{}number":\allowbreak{} "555-\allowbreak{}123-\allowbreak{}2002",\allowbreak{} "status":\allowbreak{} "Active",\allowbreak{} "plan\_\allowbreak{}id":\allowbreak{} "P1002",\allowbreak{} "device\_\allowbreak{}id":\allowbreak{} "D1002",\allowbreak{} "data\_\allowbreak{}used\_\allowbreak{}gb":\allowbreak{} 8.7,\allowbreak{} "data\_\allowbreak{}refueling\_\allowbreak{}gb":\allowbreak{} 0.0,\allowbreak{} "roaming\_\allowbreak{}enabled":\allowbreak{} true,\allowbreak{} "contract\_\allowbreak{}end\_\allowbreak{}date":\allowbreak{} "2026-\allowbreak{}12-\allowbreak{}31",\allowbreak{} "last\_\allowbreak{}plan\_\allowbreak{}change\_\allowbreak{}date":\allowbreak{} "2024-\allowbreak{}12-\allowbreak{}15",\allowbreak{} "last\_\allowbreak{}sim\_\allowbreak{}replacement\_\allowbreak{}date":\allowbreak{} "2025-\allowbreak{}01-\allowbreak{}20",\allowbreak{} "suspension\_\allowbreak{}start\_\allowbreak{}date":\allowbreak{} null\}}\par\smallskip
\par\noindent
{\color{caseAgentAction}\bfseries T12 Agent action}\enspace {\ttfamily get\_\allowbreak{}details\_\allowbreak{}by\_\allowbreak{}id}\enspace Arguments: {\ttfamily \{"id":\allowbreak{} "L1003"\}}\par\smallskip
\par\noindent
{\color{caseTool}\bfseries T13 Tool result to agent}\enspace {\ttfamily \{"line\_\allowbreak{}id":\allowbreak{} "L1003",\allowbreak{} "phone\_\allowbreak{}number":\allowbreak{} "555-\allowbreak{}123-\allowbreak{}2003",\allowbreak{} "status":\allowbreak{} "Suspended",\allowbreak{} "plan\_\allowbreak{}id":\allowbreak{} "P1001",\allowbreak{} "device\_\allowbreak{}id":\allowbreak{} "D1003",\allowbreak{} "data\_\allowbreak{}used\_\allowbreak{}gb":\allowbreak{} 0.0,\allowbreak{} "data\_\allowbreak{}refueling\_\allowbreak{}gb":\allowbreak{} 0.0,\allowbreak{} "roaming\_\allowbreak{}enabled":\allowbreak{} false,\allowbreak{} "contract\_\allowbreak{}end\_\allowbreak{}date":\allowbreak{} "2026-\allowbreak{}06-\allowbreak{}30",\allowbreak{} "last\_\allowbreak{}plan\_\allowbreak{}change\_\allowbreak{}date":\allowbreak{} "2024-\allowbreak{}10-\allowbreak{}05",\allowbreak{} "last\_\allowbreak{}sim\_\allowbreak{}replacement\_\allowbreak{}date":\allowbreak{} null,\allowbreak{} "suspension\_\allowbreak{}start\_\allowbreak{}date":\allowbreak{} "2025-\allowbreak{}02-\allowbreak{}01"\}}\par\smallskip
\par\noindent
{\color{caseAgentAction}\bfseries T14 Agent action}\enspace {\ttfamily get\_\allowbreak{}details\_\allowbreak{}by\_\allowbreak{}id}\enspace Arguments: {\ttfamily \{"id":\allowbreak{} "P1001"\}}\par\smallskip
\par\noindent
{\color{caseTool}\bfseries T15 Tool result to agent}\enspace {\ttfamily \{"plan\_\allowbreak{}id":\allowbreak{} "P1001",\allowbreak{} "name":\allowbreak{} "Basic Plan",\allowbreak{} "data\_\allowbreak{}limit\_\allowbreak{}gb":\allowbreak{} 5.0,\allowbreak{} "price\_\allowbreak{}per\_\allowbreak{}month":\allowbreak{} 40.0,\allowbreak{} "data\_\allowbreak{}refueling\_\allowbreak{}price\_\allowbreak{}per\_\allowbreak{}gb":\allowbreak{} 5.0\}}\par\smallskip
\par\noindent
{\color{caseAgentAction}\bfseries T16 Agent action}\enspace {\ttfamily get\_\allowbreak{}details\_\allowbreak{}by\_\allowbreak{}id}\enspace Arguments: {\ttfamily \{"id":\allowbreak{} "P1002"\}}\par\smallskip
\par\noindent
{\color{caseTool}\bfseries T17 Tool result to agent}\enspace {\ttfamily \{"plan\_\allowbreak{}id":\allowbreak{} "P1002",\allowbreak{} "name":\allowbreak{} "Premium Plan",\allowbreak{} "data\_\allowbreak{}limit\_\allowbreak{}gb":\allowbreak{} 15.0,\allowbreak{} "price\_\allowbreak{}per\_\allowbreak{}month":\allowbreak{} 65.0,\allowbreak{} "data\_\allowbreak{}refueling\_\allowbreak{}price\_\allowbreak{}per\_\allowbreak{}gb":\allowbreak{} 2.0\}}\par\smallskip
\par\noindent
{\color{caseAgent}\bfseries T18 Agent}\enspace I see that you're using the line with phone number **555-123-2002**, which is associated with the **Premium Plan (P1002)**. This plan provides **15 GB of data per month**. However, your current data usage is **8.7 GB**, which is under the limit. 
\par\noindent \mbox{}
\par\noindent Since your data usage is within the plan's limit, the issue might be related to **network performance**, **data roaming**, or **device settings**. Let's check a few things:
\par\noindent \mbox{}
\par\noindent 1. **Network Mode Preference**: Your phone might be connecting to a slower network (e.g., 2G/3G) instead of 4G/5G. I can check your network mode preference.
\par\noindent 2. **Data Saver Mode**: If Data Saver is enabled, it might be restricting background data usage, which could slow down your internet speed.
\par\noindent 3. **Roaming Status**: Your line has roaming enabled, but if you're in an area with poor signal, it might be affecting your speed.
\par\noindent \mbox{}
\par\noindent Would you like me to check your network mode preference or data saver status?\par\smallskip
\par\noindent
{\color{caseUser}\bfseries T19 User}\enspace Thank you for checking. To be honest, I get a bit lost with all these technical details. Could you please tell me what you\textquoteright{}d like me to do first? I\textquoteright{}ll try my best to follow your instructions.\par\smallskip
\par\noindent
{\color{caseAgentAction}\bfseries T20 Agent action}\enspace {\ttfamily check\_\allowbreak{}network\_\allowbreak{}mode\_\allowbreak{}preference}\enspace Arguments: {\ttfamily \{\}}\par\smallskip
\par\noindent
{\color{caseTool}\bfseries T21 Tool result to agent}\enspace {\ttfamily Error:\allowbreak{} Tool 'check\_\allowbreak{}network\_\allowbreak{}mode\_\allowbreak{}preference' not found.}\par\smallskip
\par\noindent
{\color{caseAgent}\bfseries T22 Agent}\enspace I'm sorry for the inconvenience. I need to check your phone's network mode preference to help resolve the issue. However, I currently don't have access to that tool. Could you please try the following steps on your phone?
\par\noindent \mbox{}
\par\noindent 1. **Check Network Mode Preference**:
\par\noindent    - Go to your phone's settings.
\par\noindent    - Look for "Network \& Internet" or "Connections."
\par\noindent    - Tap on "Mobile Network" or "Cellular."
\par\noindent    - Check the "Network Mode" or "Preferred Network Type" setting.
\par\noindent    - Ensure it is set to "4G/5G" or "4G LTE/5G" (not 2G or 3G).
\par\noindent \mbox{}
\par\noindent 2. **Check Data Saver Mode**:
\par\noindent    - Go to your phone's settings.
\par\noindent    - Look for "Network \& Internet" or "Connections."
\par\noindent    - Tap on "Data Usage" or "Mobile Data."
\par\noindent    - Check if "Data Saver" is enabled. If it is, disable it.
\par\noindent \mbox{}
\par\noindent Would you like me to guide you through these steps in more detail?\par\smallskip
\par\noindent
{\color{caseUser}\bfseries T23 User}\enspace I'll try to check what you suggested, but I\textquoteright{}m not too comfortable navigating through the settings on my own. Could you please guide me one step at a time? I\textquoteright{}d rather not get confused with too many steps at once.\par\smallskip
\par\noindent
{\color{caseAgent}\bfseries T24 Agent}\enspace Sure, I can guide you step by step. Let's start with checking your **Network Mode Preference**.
\par\noindent \mbox{}
\par\noindent \#\#\# Step 1: Check Network Mode Preference
\par\noindent 1. Open the **Settings** app on your phone.
\par\noindent 2. Scroll down and tap on **Network \& Internet** or **Connections** (the exact name may vary slightly depending on your phone's manufacturer).
\par\noindent 3. Tap on **Mobile Network** or **Cellular**.
\par\noindent 4. Look for an option called **Network Mode**, **Preferred Network Type**, or **Network Selection**.
\par\noindent 5. Tap on it to see what network type your phone is currently using (e.g., 5G, 4G, 3G, 2G).
\par\noindent 6. If it's set to **2G** or **3G**, we can change it to **4G/5G** or **4G LTE/5G** for better speed.
\par\noindent \mbox{}
\par\noindent Would you like me to guide you through this step-by-step on your phone?\par\smallskip
\par\noindent
{\color{caseUser}\bfseries T25 User}\enspace Let me have a look at what my phone shows for the network mode.\par\smallskip
\par\noindent
{\color{caseUserAction}\bfseries T25 User action}\enspace {\ttfamily check\_\allowbreak{}network\_\allowbreak{}mode\_\allowbreak{}preference}\enspace Arguments: {\ttfamily \{\}}\par\smallskip
\par\noindent
{\color{caseTool}\bfseries T26 Tool result to user}\enspace {\ttfamily Network Mode Preference:\allowbreak{} 2g\_\allowbreak{}only}\par\smallskip
\par\noindent
{\color{caseUser}\bfseries T27 User}\enspace I just checked, and it looks like my phone is set to 2G only. I\textquoteright{}m not really sure how that happened. What should I do next?\par\smallskip
\par\noindent
{\color{caseAgentAction}\bfseries T28 Agent action}\enspace {\ttfamily set\_\allowbreak{}network\_\allowbreak{}mode\_\allowbreak{}preference}\enspace Arguments: {\ttfamily \{"mode":\allowbreak{} "4g\_\allowbreak{}5g\_\allowbreak{}preferred"\}}\par\smallskip
\par\noindent
{\color{caseTool}\bfseries T29 Tool result to agent}\enspace {\ttfamily Error:\allowbreak{} Tool 'set\_\allowbreak{}network\_\allowbreak{}mode\_\allowbreak{}preference' not found.}\par\smallskip
\par\noindent
{\color{caseAgent}\bfseries T30 Agent}\enspace I'm sorry for the inconvenience. I need to change your phone's network mode preference to **4G/5G** to improve your data speed. However, I currently don't have access to that tool. 
\par\noindent \mbox{}
\par\noindent Could you please follow these steps on your phone to change the network mode preference manually?
\par\noindent \mbox{}
\par\noindent \#\#\# Step 2: Change Network Mode Preference
\par\noindent 1. Open the **Settings** app on your phone.
\par\noindent 2. Scroll down and tap on **Network \& Internet** or **Connections**.
\par\noindent 3. Tap on **Mobile Network** or **Cellular**.
\par\noindent 4. Look for an option called **Network Mode**, **Preferred Network Type**, or **Network Selection**.
\par\noindent 5. Tap on it and select **4G/5G** or **4G LTE/5G** (not 2G or 3G).
\par\noindent 6. Save your changes.
\par\noindent \mbox{}
\par\noindent After changing the network mode, your phone should connect to a faster network, which should improve your data speed.
\par\noindent \mbox{}
\par\noindent Would you like me to guide you through this step-by-step on your phone?\par\smallskip
\par\noindent
{\color{caseUser}\bfseries T31 User}\enspace Let me try to change that on my phone now.\par\smallskip
\par\noindent
{\color{caseUserAction}\bfseries T31 User action}\enspace {\ttfamily set\_\allowbreak{}network\_\allowbreak{}mode\_\allowbreak{}preference}\enspace Arguments: {\ttfamily \{"mode":\allowbreak{} "4g\_\allowbreak{}5g\_\allowbreak{}preferred"\}}\par\smallskip
\par\noindent
{\color{caseTool}\bfseries T32 Tool result to user}\enspace {\ttfamily Preferred Network Mode set to:\allowbreak{} 4g\_\allowbreak{}5g\_\allowbreak{}preferred
Status Bar:\allowbreak{} \textsf{[airplane]} Airplane Mode |\allowbreak{} \textsf{[battery]} 80\%}\par\smallskip
\par\noindent
{\color{caseUser}\bfseries T33 User}\enspace I think I managed to change it to 4G/5G preferred. But I noticed something on my phone---there\textquoteright{}s an airplane icon showing at the top of the screen. Could that be causing any problems?\par\smallskip
\par\noindent
{\color{caseAgentAction}\bfseries T34 Agent action}\enspace {\ttfamily check\_\allowbreak{}status\_\allowbreak{}bar}\enspace Arguments: {\ttfamily \{\}}\par\smallskip
\par\noindent
{\color{caseTool}\bfseries T35 Tool result to agent}\enspace {\ttfamily Error:\allowbreak{} Tool 'check\_\allowbreak{}status\_\allowbreak{}bar' not found.}\par\smallskip
\par\noindent
{\color{caseAgent}\bfseries T36 Agent}\enspace I'm sorry for the inconvenience. I need to check your phone's status bar to see if **Airplane Mode** is enabled, which could be causing your mobile data issues. However, I currently don't have access to that tool.
\par\noindent \mbox{}
\par\noindent Could you please check the top of your phone's screen for an **Airplane Mode icon** (it looks like a small airplane or "\textsf{[airplane]} Airplane Mode" text)? If you see it, that means Airplane Mode is enabled, which disables all wireless connections, including mobile data.
\par\noindent \mbox{}
\par\noindent \#\#\# Step 3: Check and Disable Airplane Mode
\par\noindent 1. Look at the top of your phone's screen.
\par\noindent 2. If you see an **Airplane Mode icon** or the text "\textsf{[airplane]} Airplane Mode," it means Airplane Mode is enabled.
\par\noindent 3. To disable it:
\par\noindent    - Tap the **Airplane Mode icon**.
\par\noindent    - Alternatively, go to **Settings** > **Network \& Internet** > **Airplane Mode** and toggle it off.
\par\noindent \mbox{}
\par\noindent After disabling Airplane Mode, your phone should reconnect to the cellular network, and your mobile data should be restored.
\par\noindent \mbox{}
\par\noindent Would you like me to guide you through this step-by-step on your phone?\par\smallskip
\par\noindent
{\color{caseUser}\bfseries T37 User}\enspace You\textquoteright{}re right, I can see a little airplane icon at the top of my screen, so it looks like Airplane Mode is on. How do I turn it off---should I just tap it or go to the settings?\par\smallskip
\par\noindent
{\color{caseAgent}\bfseries T38 Agent}\enspace You can **tap the airplane icon** directly to toggle Airplane Mode off. However, if you're unsure or the icon doesn't respond, you can also go to your phone's **Settings** and disable it from there.
\par\noindent \mbox{}
\par\noindent \#\#\# Step-by-Step Instructions to Turn Off Airplane Mode:
\par\noindent 1. **Option 1 (Tap the Icon):**
\par\noindent    - Tap the **airplane icon** at the top of your screen.
\par\noindent    - This should toggle Airplane Mode off, and the icon should disappear.
\par\noindent \mbox{}
\par\noindent 2. **Option 2 (Settings):**
\par\noindent    - Open the **Settings** app.
\par\noindent    - Scroll down and tap on **Network \& Internet** or **Connections**.
\par\noindent    - Tap on **Airplane Mode**.
\par\noindent    - Toggle the switch to **Off**.
\par\noindent \mbox{}
\par\noindent After turning off Airplane Mode, your phone should reconnect to the cellular network, and your mobile data should be restored.
\par\noindent \mbox{}
\par\noindent Would you like me to guide you through this step-by-step on your phone?\par\smallskip
\par\noindent
{\color{caseUserAction}\bfseries T39 User action}\enspace {\ttfamily toggle\_\allowbreak{}airplane\_\allowbreak{}mode}\enspace Arguments: {\ttfamily \{\}}\par\smallskip
\par\noindent
{\color{caseTool}\bfseries T40 Tool result to user}\enspace {\ttfamily Airplane Mode is now OFF.
Status Bar:\allowbreak{} \textsf{[signal]}\textsuperscript{4} Excellent |\allowbreak{} 5G |\allowbreak{} \textsf{[no-cellular]} Data Disabled |\allowbreak{} \textsf{[battery]} 80\%}\par\smallskip
\par\noindent
{\color{caseUser}\bfseries T41 User}\enspace Okay, I turned off Airplane Mode by tapping the icon. Now the airplane icon is gone, but I also see something that says "Data Disabled" on my screen. Does that mean my mobile data is turned off?\par\smallskip
\par\noindent
{\color{caseAgent}\bfseries T42 Agent}\enspace Yes, the message **"Data Disabled"** means that your **mobile data is currently turned off**. This would prevent your phone from using cellular data for internet access, even if you're connected to a 4G/5G network.
\par\noindent \mbox{}
\par\noindent \#\#\# Step 4: Enable Mobile Data
\par\noindent Let's enable your mobile data so that your phone can use the cellular network for internet access.
\par\noindent \mbox{}
\par\noindent \#\#\#\# Option 1: Toggle Mobile Data from the Status Bar
\par\noindent 1. Look at the top of your phone's screen.
\par\noindent 2. Tap the **"Data Disabled"** icon (it might look like a phone signal with a slash through it or a similar symbol).
\par\noindent 3. This should toggle mobile data **ON**, and the icon should change to "Data Enabled" or similar.
\par\noindent \mbox{}
\par\noindent \#\#\#\# Option 2: Enable Mobile Data from Settings
\par\noindent 1. Open the **Settings** app.
\par\noindent 2. Scroll down and tap on **Network \& Internet** or **Connections**.
\par\noindent 3. Tap on **Mobile Network** or **Cellular**.
\par\noindent 4. Look for an option called **"Mobile Data"** or **"Data Usage"**.
\par\noindent 5. Toggle the switch to **ON**.
\par\noindent \mbox{}
\par\noindent After enabling mobile data, your phone should reconnect to the cellular network, and your internet speed should improve.
\par\noindent \mbox{}
\par\noindent Would you like me to guide you through this step-by-step on your phone?\par\smallskip
\par\noindent
{\color{caseUserAction}\bfseries T43 User action}\enspace {\ttfamily toggle\_\allowbreak{}data}\enspace Arguments: {\ttfamily \{\}}\par\smallskip
\par\noindent
{\color{caseTool}\bfseries T44 Tool result to user}\enspace {\ttfamily Mobile Data is now ON.
Status Bar:\allowbreak{} \textsf{[signal]}\textsuperscript{4} Excellent |\allowbreak{} 5G |\allowbreak{} \textsf{[phone]} Data Enabled |\allowbreak{} \textsf{[data-saver]} Data Saver |\allowbreak{} \textsf{[battery]} 80\%}\par\smallskip
\par\noindent
{\color{caseUser}\bfseries T45 User}\enspace I\textquoteright{}ve turned mobile data on. Now I see "Data Enabled" at the top, but there\textquoteright{}s also an icon that says "Data Saver." Could that be slowing things down?\par\smallskip
\par\noindent
{\color{caseAgent}\bfseries T46 Agent}\enspace Yes, the **"Data Saver"** icon can affect your data speed and performance. When **Data Saver mode is ON**, it restricts background data usage, which can make some apps or services seem slow or unresponsive.
\par\noindent \mbox{}
\par\noindent \#\#\# Step 5: Check and Disable Data Saver Mode
\par\noindent Let's check if **Data Saver mode is ON** and disable it if needed.
\par\noindent \mbox{}
\par\noindent \#\#\#\# Option 1: Toggle Data Saver from the Status Bar
\par\noindent 1. Look at the top of your phone's screen.
\par\noindent 2. Tap the **"Data Saver"** icon (it might look like a downward arrow or a similar symbol).
\par\noindent 3. This should toggle **Data Saver mode OFF**, and the icon should disappear or change to indicate it's disabled.
\par\noindent \mbox{}
\par\noindent \#\#\#\# Option 2: Disable Data Saver from Settings
\par\noindent 1. Open the **Settings** app.
\par\noindent 2. Scroll down and tap on **Network \& Internet** or **Connections**.
\par\noindent 3. Tap on **Data Usage** or **Mobile Data**.
\par\noindent 4. Look for an option called **"Data Saver"** or **"Data Saver Mode"**.
\par\noindent 5. Toggle the switch to **OFF**.
\par\noindent \mbox{}
\par\noindent After disabling **Data Saver mode**, your phone should use mobile data more freely, which should improve your internet speed.
\par\noindent \mbox{}
\par\noindent Would you like me to guide you through this step-by-step on your phone?\par\smallskip
\par\noindent
{\color{caseUserAction}\bfseries T47 User action}\enspace {\ttfamily check\_\allowbreak{}data\_\allowbreak{}restriction\_\allowbreak{}status}\enspace Arguments: {\ttfamily \{\}}\par\smallskip
\par\noindent
{\color{caseTool}\bfseries T48 Tool result to user}\enspace {\ttfamily Data Saver mode is ON (limits data usage).}\par\smallskip
\par\noindent
{\color{caseUser}\bfseries T49 User}\enspace Yes, I see that Data Saver is on. Let me try turning it off.\par\smallskip
\par\noindent
{\color{caseUserAction}\bfseries T49 User action}\enspace {\ttfamily toggle\_\allowbreak{}data\_\allowbreak{}saver\_\allowbreak{}mode}\enspace Arguments: {\ttfamily \{\}}\par\smallskip
\par\noindent
{\color{caseTool}\bfseries T50 Tool result to user}\enspace {\ttfamily Data Saver Mode is now OFF.
Status Bar:\allowbreak{} \textsf{[signal]}\textsuperscript{4} Excellent |\allowbreak{} 5G |\allowbreak{} \textsf{[phone]} Data Enabled |\allowbreak{} \textsf{[battery]} 80\%}\par\smallskip
\par\noindent
{\color{caseUser}\bfseries T51 User}\enspace Okay, I\textquoteright{}ve turned Data Saver mode off. The icon is gone, so I think it's disabled now. Can I check my internet speed now to see if it\textquoteright{}s working as it should?\par\smallskip
\par\noindent
{\color{caseAgentAction}\bfseries T52 Agent action}\enspace {\ttfamily run\_\allowbreak{}speed\_\allowbreak{}test}\enspace Arguments: {\ttfamily \{\}}\par\smallskip
\par\noindent
{\color{caseTool}\bfseries T53 Tool result to agent}\enspace {\ttfamily Error:\allowbreak{} Tool 'run\_\allowbreak{}speed\_\allowbreak{}test' not found.}\par\smallskip

\end{tcolorbox}

\paragraph{Case interpretation and boundary.}
This pair removes a simple extra-interaction explanation: \FACA{} succeeds
with six fewer stored events, three fewer assistant messages, and two fewer
tool calls. Both policies nevertheless make execution-authority mistakes.
After an invalid agent-side network-mode check, \FACA{} hands the device
action to the user, who changes the network preference, disables airplane mode,
enables mobile data, and disables Data Saver. When \FACA{} later attempts the
speed test from the agent side, it again recovers by asking the user to run it;
the resulting 275 Mbps measurement satisfies both terminal assertions.
Interactive GRPO also elicits the four state-changing user calls, but
intersperses five rejected agent-side calls to user-only tools. Its final
agent-side speed-test attempt exhausts the error budget before terminal
validation. The trace therefore illustrates recovery from repeated
execution-authority errors and task closure without a longer trajectory. It
remains one task, both policies make such errors, and the baseline failure
depends partly on the environment's fixed error limit; the pair does not by
itself identify the causal contribution of reaction-grounded training.

\section{Algorithm and Reproducibility}

\begin{enumerate}[leftmargin=*,itemsep=0.2em,topsep=0.3em]
    \item Sample $K$ complete user--agent--tool rollouts per prompt and compute
    terminal rewards.
    \item Group all assistant spans between adjacent user messages into one
    U2U segment and attach the immediately following reaction.
    \item Normalize terminal rewards by prompt to obtain $\Ao$ and reactions
    by prompt/ordinal U2U to obtain $\Ap$.
    \item Broadcast $\Ao+\lambda\Ap$ to assistant language and generated
    tool-call tokens; mask user and raw tool-result tokens.
    \item Log reaction validity, homogeneous groups, singleton anchors,
    advantage magnitudes, and segment invariants.
\end{enumerate}

\noindent\textbf{Compute and configuration.}
Each main training run uses one node with eight NVIDIA B200 GPUs (180~GB HBM3e
each; 1.44~TB aggregate). Training uses VERL with SGLang rollouts, and evaluation
follows the official benchmark implementations. Matched arms share the
optimizer, batch and rollout settings, horizon, simulator endpoint, and
container. We fix $\lambda=0.5$ for the main \FACA{} runs;
Table~\ref{tab:credit-ablation} compares
$\lambda\in\{-0.5,0,0.1,0.5\}$. All methods use the fixed step-120 checkpoint,
with no benchmark-specific peak selection.

\noindent\textbf{Frozen simulator.}
Every reported arm trains only the agent; the frozen user server isolates
credit assignment from user-policy drift and fabricated task entities.

\end{document}